\documentclass{article}

\PassOptionsToPackage{numbers, compress}{natbib}
\usepackage[preprint]{neurips_2026}

\usepackage[utf8]{inputenc} 
\usepackage[T1]{fontenc}    
\usepackage{hyperref}       
\usepackage{url}            
\usepackage{booktabs}       
\usepackage{amsfonts}       
\usepackage{nicefrac}       
\usepackage{microtype}      
\usepackage{xcolor}         
\usepackage{amsmath}
\usepackage{graphicx}
\usepackage{booktabs}
\usepackage{multirow}
\usepackage{rotating}
\usepackage[table]{xcolor}
\usepackage{colortbl}
\usepackage{wrapfig}
\definecolor{gold}{rgb}{1.0, 0.84, 0.0}
\definecolor{H10color}{HTML}{37B221}
\definecolor{C11color}{HTML}{4A5568}
\definecolor{N13color}{HTML}{047DD9}
\definecolor{O10color}{HTML}{FF6363}
\definecolor{Hcolor}{HTML}{FFFFFF}
\definecolor{Ccolor}{HTML}{B4B4B4}
\definecolor{Ncolor}{HTML}{778EFE}
\definecolor{Ocolor}{HTML}{FF0D0D}
\newcommand{\Cv}{\ensuremath{C_{\mathrm{v}}}}

\title{QT-Net: Rethinking Evaluation of AI Models in Atomic Chemical Space}

\author{Pablo Martínez Crespo \\
  Department of Computer Science and Engineering\\
  Chalmers University of Technology
  and University of Gothenburg\\
  \texttt{pabloma@chalmers.se} \\
  \And
  Stefano Ribes \\
  Department of Computer Science and Engineering\\
  Chalmers University of Technology
  and University of Gothenburg\\
  \And
  Richard Beckmann \\
  Department of Computer Science and Engineering\\
  Chalmers University of Technology
  and University of Gothenburg\\
  \And
  Marisa Gliege \\
  Chief Technology Office \\
  EMD Electronics \\
  \And
  Santiago Miret \\
  Lila Sciences \\
  \And
  Robert S. Jordan \\
  Technology Research \\
  Intel Corporation \\
  \And
  Vijay Kris Narasimhan \\
  M Ventures \\
  \And
  Martin Rahm \\
  Department of Chemistry and Chemical Engineering\\
  Chalmers University of Technology\\
  \And
  Rocío Mercado \\
  Department of Computer Science and Engineering\\
  Chalmers University of Technology and University of Gothenburg\\
  and Science for Life Laboratory (SciLifeLab), Gothenburg \\
  \texttt{rocio.mercado@chalmers.se} \\
}

\begin{document}

\maketitle

\begin{abstract}
  Atomic properties such as partial charges or multipoles encode chemically meaningful information that can inform downstream molecular property prediction, but their evaluation as machine learning targets has been complicated by the absence of a principled out-of-distribution evaluation protocol at the atomic level. In this work, we propose a held-out evaluation protocol that clusters atomic environments by SOAP descriptors and computes metrics accounting only for cluster labels unseen during training. Following this procedure, we use 5×5 cross-validation and Tukey's HSD to run a statistically rigorous comparison of E(3)-equivariant against non-equivariant, rotationally augmented models for predicting electron populations and multipoles of H, C, N, and O atoms. Building on our results, we introduce the Quantum Topological Neural Network (QT-Net), a rotationally augmented, non-equivariant graph neural network. We show that QT-Net can be used to infer properties of atoms in molecules from QM9 outside our training set, and that the atomic environments seen during training strongly affect the performance of molecular property prediction tasks. To further validate the framework, molecular dipole moments computed from QT-Net's per-atom outputs recover the ground-truth values reported in QM9. We release all code and data, including a JAX implementation of QT-Net.
\end{abstract}

\section{Introduction}

Molecular property prediction \cite{levine2025open, passaro2025boltz}, accelerated simulations \cite{kovacs2025mace, lewis2025scalable}, and \textit{de novo} molecular design \cite{zeni2025generative, he2026democratising} are among the main drivers of new artificial intelligence (AI) for science methods. However, out-of-distribution (OOD) generalization and broad sampling are among the most pressing challenges in the field.~\cite{gao2022sample, vskrinjar2026evaluating, antoniuk2026boom} Molecular representations built from first principles atomic properties are a promising avenue for tackling this research gap.~\cite{jorner2021machine, stuyver2022quantum, guan2021regio, li2024quantum, alfonso2024repurposing}

Because atoms in molecules and materials interact as open quantum systems that exchange energy and charge with their neighbors, not all atoms of the same element can be treated as equal. Quantum chemists have developed several atomic descriptors that can also be effective for downstream property prediction, including conceptual density functional theory (DFT) reactivity indices such as Fukui functions~\cite{parr1984density}, electrostatic surface potentials and local ionization energies~\cite{jorner2021machine}, natural population analysis~\cite{stuyver2022quantum}, interacting quantum atoms (IQA) energy decompositions~\cite{maxwell2016prediction,brown2024incorporating,hoffmann2026quantum}, and the various definitions of atomic partial charge~\cite{metcalf2020electron,wang2021deepatomiccharge,bereau2015transferable,gallegos2022nnaimq}. Works by \citet{vargas2024high} and \citet{gee2025multi} showed that descriptors derived from the quantum theory of atoms in molecules (QTAIM) can enhance regression tasks such as predicting activation energies in reactions, or formation and orbital energies of transition metal complexes with less than 1\,000 training points. Recently, the AIMEl dataset ~\cite{meza2024quantum}, a subset of QM9 \cite{ramakrishnan2014quantum} with quantum topological atomic (QTA) properties for H, C, N and O such as electron populations, multipole moments, and localization indices, was released. Some open questions from these works are whether QTA properties inferred by a model would preserve their predictive power, and which properties are the most informative outside the region of chemical space that a model was trained on.  

Several studies have focused on learning atomic partial charges and multipoles derived from first-principles.~\cite{metcalf2020electron,kato2020high,wang2021deepatomiccharge,veit2020predicting,bereau2015transferable} Nevertheless, truly OOD evaluation is precluded by the usage of random splits, which cannot prevent atomic environments from leaking into test sets. Atoms, as the building blocks of chemistry and molecules, are open quantum systems with complex interactions of their own dictated by their electronic structure. In this work, we show that metrics that account for all atoms in a test set of molecules, rather than only atoms in environments unseen during training, lead to overconfident results that don't reflect true OOD performance. 

In view of this, we introduce the Quantum Topological Neural Network (QT-Net) for learning QTA properties, a densely connected, non-equivariant graph neural network (GNN). To address the aforementioned limitations, our key contributions are:
\begin{itemize}
    \item \textbf{Statistically significant benchmarking.} We apply repeated-measures ANOVA (RM-ANOVA) and Tukey's honestly significant difference (HSD) frameworks for comparing E(3)-equivariant against non-equivariant multitask models to predict scalar and tensor properties for topological atoms.
    \item \textbf{Faithful OOD evaluation.} Our evaluation set, built upon per-element clustering of smooth overlap of atomic positions (SOAP), contains atomic environments unseen during training, and our metrics only account for atoms in said unseen environments.
    \item \textbf{Inferential quality assessment.} We infer QTA properties for all the QM9 atoms outside of AIMEl, and show that the molecular dipole moment is recovered from its atomic contributions.
\end{itemize}

\section{Related work}\label{sec:related}

A productive line of work has applied machine learning (ML) to QTA properties drawn from QTAIM, the IQA formalism, and related branches of quantum chemical topology (QCT), establishing that these chemical descriptors are within reach of modern regressors. Early contributions used Gaussian process regression to predict the full set of IQA energy components for small biomolecular fragments ~\cite{maxwell2016prediction}, and were subsequently extended to clusters with the formamide-dimer transfer-learning study of \citet{brown2024incorporating}, which expanded the property catalog to atomic multipole moments. The works by Gallegos et al. \cite{gallegos2022nnaimq,gallegos2024explainable} introduced neural networks for predicting QTAIM atomic charges of H, C, N, and O atoms, and one-body and two-body QTAIM and IQA descriptors. Most recently, \citet{hoffmann2026quantum} have shown that equivariant graph networks can learn IQA-decomposed atomic and pairwise energies across diverse benchmark sets. Together, these works establish QTA properties as a realistic ML target and have steadily expanded the space of accessible descriptors.

A methodological pitfall nevertheless recurs across this literature, bearing directly on the OOD-evaluation contribution we make in this work. Mainly, these works rely on splits that do not isolate unseen atomic environments. \citet{maxwell2016prediction} and \citet{gallegos2022nnaimq} draw their splits at random from a common pool of geometries, which is bound to leak atomic environments seen during training into the test set. Their follow-up \cite{gallegos2024explainable} probes extrapolation along reaction coordinates and through a supramolecular trajectory — useful as configurational-space stress tests, but uninformative about generalization across chemical space. \citet{brown2024incorporating} sample training and evaluation from a single molecular-dynamics trajectory of the same monomer. They show that their monomer-trained multipoles fail to transfer to the dimer regime, providing perhaps the clearest signal that environment-level OOD is a standing challenge.

\section{Methods}

Figure~\ref{fig:pipeline} summarizes the methodology followed to train and evaluate the proposed QT-Net models.

\subsection{Target properties of topological atoms}
QTAIM defines an atom as a region in real space $\Omega\subset\mathbb{R}^3$ such that $\nabla \rho (\mathbf{r}) \cdot \mathbf{\hat{n}}(\mathbf{r}) = 0\hspace{1mm}\forall \ \mathbf{r}\in\partial\Omega$,
where $\rho$ is the electron density and $\mathbf{\hat{n}}(\mathbf{r})$ is the normal vector to the surface defined by $\partial\Omega$, meaning there is no flux of electrons through the boundary of the basin. Atomic properties depend on the shape of this basin, as they are quantities integrated over it. In this work, we focus on electron population ($N$), atomic contribution to molecular dipole moment ($\boldsymbol{\mu}$), traceless quadrupole moment ($\mathbf{Q}$), and localization index ($\lambda$),
\begin{equation*}
    N=\int_\Omega \rho(\mathbf{r})d\mathbf{r},
\quad
\boldsymbol{\mu} = -\int_\Omega e\,\mathbf{r}^{(\Omega)}\rho(\mathbf{r})d\mathbf{r}+
    \sum_\Lambda^{N_b(\Omega)}\left[\mathbf{R}_\Omega-\mathbf{R}_b(\Omega |\Lambda)\right]q(\Omega|\Lambda),
\end{equation*}
\begin{equation*}
    \text{and} \ Q_{\alpha\beta} = -\int_\Omega e\,\rho(\mathbf{r}) \left(3r_\alpha^{(\Omega)}r_\beta^{(\Omega)} -\delta_{\alpha\beta} \|\mathbf{r}^{(\Omega)} \|^2\right)d\mathbf{r}.
\end{equation*}
In the expressions above, $\rho(\mathbf{r})$ is the electron density, $e$ is the charge of the electron, indices $\alpha,\beta \in \{x,y,z\}$ denote Cartesian components s.t. $r_{\alpha}^{(\Omega)}$ is the $\alpha$ component of the displacement from the nuclear position $\mathbf{R}_\Omega$, $N_b(\Omega)$ is the number of bond critical points (BCPs) connected to atom $\Omega$, $\mathbf{R}_b(\Omega|\Lambda)$ is the position of the BCP between atoms $\Omega$ and $\Lambda$ and $q(\Omega|\Lambda)$ is the charge transfer between these atoms. The localization index $\lambda$ is defined from the variance of the electron population within a basin as $\lambda = N-\sigma^2(N)$, and quantifies the extent to which electrons are localized within the basin $\Omega$ vs shared with neighboring basins. We refer the interested readers to the book by \citet{matta2007introduction}. Here $N$ is a scalar, $\mu$ is a 3-vector, $Q$ is a rank-2 traceless symmetric tensor with 5 independent components, and $\lambda$ is a scalar.

\subsection{Dataset construction and atomic environments}\label{sec:dataset}

\begin{figure}
    \centering
    \includegraphics[width=\linewidth]{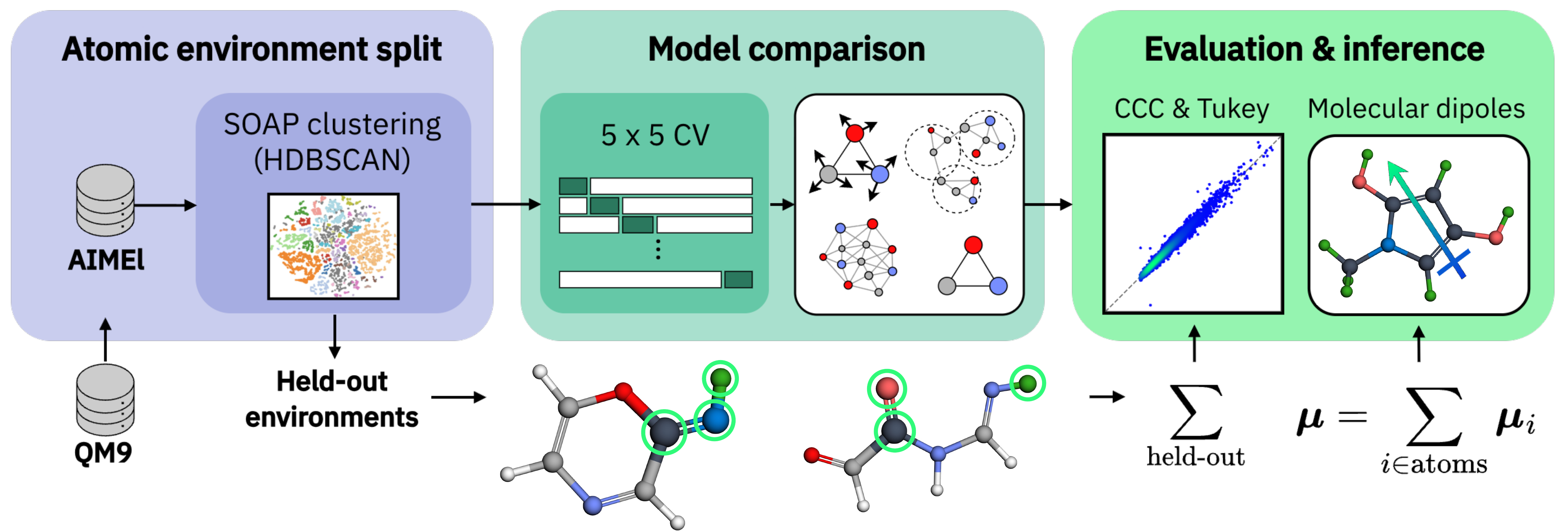}
    \caption{QT-Net training and evaluation pipeline on AIMEl for atomic (QTAIM) property prediction. The molecules at the bottom are sample molecules from the test set (left: \texttt{N=C1OC=CN=C1}; right: \texttt{N=CNC(=O)C=O}), where the circled atoms are in environments held-out from training and are the only ones contributing to the scores. The molecule with the dipole moment (\texttt{CN1C=C(O)C=C1O}) is part of the physics-based test outside AIMEl. Color for atoms in held-out environments: \colorbox{H10color}{\textcolor{white}{H\_10}},
\colorbox{C11color}{\textcolor{white}{C\_11}},
\colorbox{N13color}{\textcolor{white}{N\_13}},
\colorbox{O10color}{O\_10}. Color for the other atoms: 
\colorbox{Hcolor}{H},
\colorbox{Ccolor}{C},
\colorbox{Ncolor}{\textcolor{white}{N}},
\colorbox{Ocolor}{O}.} \vspace{-.4cm}
    \label{fig:pipeline}
\end{figure}

We used the QTAIM atomic multipoles ($N$, $\boldsymbol{\mu}$, $\mathbf{Q}$) and localization indices ($\lambda$) reported in AIMEl \cite{meza2024quantum}. The AIMEl Zenodo repository \cite{meza2024zenodo} releases data in two formats: a summary \texttt{.csv} file covering approximately 40\,000 molecules but containing only scalar properties, and per-molecule \texttt{.sumviz} files (raw AIMAll output) that include the full tensor components of $\boldsymbol{\mu}$ and $\mathbf{Q}$. To enable the comparison of equivariant and non-equivariant models on tensorial targets, we parsed the \texttt{.sumviz} files directly. These were available for 31\,223 of the molecules, of which we removed 411 entries with missing molecular property values. The resulting AIMEl-derived dataset contains 30\,812 molecules.

To enable evaluation on atomic environments not seen during training, we group atoms into element-specific clusters of similar local environment, which are later used to construct held-out test sets of atoms whose cluster labels never appear in the training data. For each atom we computed the SOAP descriptors \cite{bartok2013representing} (cutoff 4.5\,Bohr, 8 radial basis functions, spherical harmonics up to $L=6$, Gaussian smearing $\sigma = 0.7$\,Bohr; 3696 features per atom). These were reduced using principal component analysis (PCA) to element-specific numbers of features preserving at least 99\% of the variance, and clustered separately by atomic species using HDBSCAN \cite{mcinnes2017hdbscan}. Each atom in the dataset was then annotated with its element-centered environment label. Cluster statistics are reported in Table~\ref{tab:clustering} and Figures \ref{fig:atom_label_count} and \ref{fig:atom_label_spread} in the Appendix. 

We now describe the co-occurrence closure construction of the held-out set for the evaluation of QTA models. For reproducibility purposes, we use A\_$i$ in the following to refer to the label of an atomic environment centered around element A, as these are the labels reported with our dataset. We selected N\_13 first, as nitrogen was the limiting factor for holdout size; over 90\% of molecules containing N\_13 also contained C\_11 and H\_10, so both were added to prevent co-occurrence leakage. Including these three labels automatically absorbed all O\_10-containing molecules, which we therefore retained as a fourth evaluation label at no additional cost in training-set size. The final test set contains 5\,840 molecules, with at least one unseen cluster label per atom type (see the bottom of Fig. \ref{fig:pipeline}), forming a coherent OOD region of chemical space at both the molecular and atomic levels.

\subsubsection{Data curation for molecular experiments}

Our experiments on molecular property prediction run on QM9 \cite{ramakrishnan2014quantum}, of which AIMEl is a subset. We exclude the 3\,054 molecules reported to fail the geometry-consistency check (molecules whose DFT-optimized geometry does not match their input SMILES—flagged by the QM9 authors as a known issue). Molecules containing fluorine are also excluded to enable the study of how QTA-properties affect downstream molecular tasks, as there is no data for F in AIMEl. In total, the working dataset for these experiments contains 128\,908 molecules, with 98\,096 for which QTAIM properties were inferred using QT-Net. We retain four target properties: isotropic polarizability ($\alpha$), HOMO--LUMO energy gap ($\Delta\varepsilon$), internal energy at 0\,K ($U_0$), and heat capacity at 298.15\,K (\Cv). Together, these span both electrostatic-response and energetic properties, for which the estimation of atomic multipoles is plausibly informative.

These 128\,908 molecules provide a total of 2\,333\,341 atoms which are clustered by environment following the protocol of Section \ref{sec:dataset}, except for a SOAP cutoff of 6 Bohr to match the receptive radius of the molecular models. This alternative set of atomic labels allows us to analyze the extrapolation regime of the inferred QTAIM properties respect to the training distribution in atomic chemical space, and whether atomic environments play a role in molecular property prediction. At the same time, we enable OOD evaluation in molecular chemical space by assigning each molecule to a `chemotype' cluster. For this, we compute binary Morgan fingerprints (radius 2, 2\,048 bits), normalize them to unit L2 norm, project them on 50 principal components, and cluster them with k-means (k = 50, 10 initializations, fixed seed).

\subsection{Cross-validation at the atomic and molecular levels}\label{sect:CV}

All models are evaluated using a 5-repeat, 5-fold cross-validation (5×5 CV) protocol, yielding 25 fold-level scores per model. A standard 5×5 CV protocol computes performance metrics on fold-specific holdouts drawn from the data. Here, predicting atomic properties in the data regime of the AIMEl dataset forces the usage of a common held-out set of atomic environments for all 25 folds. We justify this design in the Limitations (Sec. \ref{sec:Limitations}) and in the Appendix \ref{SI:cutoff_leakage}, and provide metrics of the statistical validity of the comparison in the Appendix \ref{app:cv_diagnostics}. The remaining AIMEl molecules are partitioned into fold-specific training and validation sets. Within each repeat, molecules are split into five folds using \texttt{GroupKFold} with the Bemis--Murcko scaffold as the grouping variable, so all molecules sharing a scaffold are assigned to the same fold; each fold's validation set therefore contains only scaffolds entirely absent from the training set. Scaffold-to-fold assignments are re-shuffled with an independent seed at each repeat, making the five repeats statistically distinct.

For the molecular experiments, we follow a standard 5×5 CV protocol drawing the folds from AIMEl and leaving the remainder of QM9 purely for inference. Each fold withholds a group of molecular chemotypes and a set of atomic environments at the same time. The 48 chemotypes present in AIMEl are dealt into five groups of whole chemotypes with balanced molecule counts. Environments are withheld through five environment folds, which automate the co-occurrence closure of Sec.~\ref{sec:dataset} on the 6-Bohr labels. Fold $(r,c)$ of repeat $r$ withholds chemotype group $c$ and environment fold $(r+c) \bmod 5$, making each chemotype group meet each environment fold once over the five repeats. Every molecule with an atom in a withheld environment leaves training, too. Because a molecule keeps its chemotype group while the withheld environment fold rotates between repeats, it is tested both by models that saw its environments and by models that did not. See Appendix \ref{SI:QM9_clustering} for further details and t-SNE visualizations.

\subsection{Statistical framework of the comparison of models for QTA properties}

We compare model performance using a repeated-measures ANOVA (RM-ANOVA) with Tukey's honestly significant difference (HSD) as the post-hoc pairwise test. Data limitations force a common hold-out set across all folds (Sec. ~\ref{sec:Limitations}), on which we compute per-fold concordance correlation coefficients (CCC) for each (holdout label, property). Because the five folds within a repeat share this hold-out and overlap in training data, their scores are correlated. We therefore average them into a single repeat-level score, yielding five approximately independent observations per model in a balanced design with repeat as subject and model as treatment. Means and 95\% confidence intervals are reported as $\bar{x} \pm t_{0.975,4} \cdot \mathrm{SEM}$, where SEM is the standard error of the five repeat-level means. Tukey HSD then defines a minimum significant difference (MSD) such that two models differ significantly only if the gap between their repeat-averaged means exceeds it. The same procedure applies to any subset of (atom $\times$ property) cells, and we report the diagnostics of normality, homoscedasticity and fold independence in Appendix~\ref{app:cv_diagnostics}.

\subsection{The QT-Net architecture}

\vspace{-10pt}
\begin{figure}[h]
    \centering
    \includegraphics[width=\linewidth]{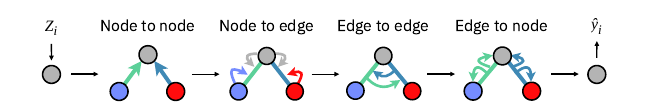}
    \caption{Schematic figure of a layer of message passing of QT-Net.}
    \label{fig:qtnet-layer}
\end{figure}

The set of transformations that compose a layer of QT-Net is represented in Figure~\ref{fig:qtnet-layer}. We focus on the transformations for the scalar versions of QT-Net, as that was the best-performing design. See Section \ref{SI:equivariant} for details on E(3)-equivariant models. Our architecture is inspired by the attention and gating mechanisms of the work by \citet{irwin2024semlaflow}, the separation of feature and geometric channels by \citet{schutt2021equivariant}, and the chemical reminders, raw-component input, and per-layer readout by \citet{pozdnyakov2023smooth}.

We represent each atom by a node with scalar features $h_i$. For edge features, we follow the same convention, but with double subindices (e.g., $h_{ij}$). We denote the element-wise (Hadamard) product with $\odot$. In the equations below that can apply to either type of cell, we absorb the subindices of nodes and edges into the subscripts $r,\ s,\ t$ to refer to receivers, senders and intermediaries (respectively). In messages between nodes, the intermediary would be the edge connecting them (and the common node between two edges for messages between edges). The usage of Hadamard product implies that the output dimension of MLPs and geometric filters match the number of features of the receiver. We use SiLU as activation function between hidden layers.

In all models, node features are initialized with a message passing layer as follows:
\begin{equation}\label{eq:node_encoder}
    h_i^{(0)} = \mathrm{LayerNorm}\left(\sum_j \mathrm{MLP}\left(\mathrm{EMBED}(Z_i),\mathrm{EMBED}(Z_j)\right) \odot \Gamma\left(\hat{\mathbf{r}}_{ij},\mathrm{RBF}(r_{ij})\right)\right),
\end{equation}
where $Z_i,Z_j$, are the atomic species. $\Gamma$ is a geometric gate with $\tanh$ activation output that takes as input the components of the inter-atom displacement vector, concatenated to a projection onto radial basis functions (RBFs) as implemented in DimeNet \cite{gasteiger2020directional}: 
\begin{equation*}
    \text{RBF}_n(r) = \sqrt{\frac{2}{c}}\frac{\sin(n\pi r/c)}{r}\frac{1+\cos(\pi r/c)}{2}.
\end{equation*}

Edge features are initialized from these node features as follows
\begin{equation}\label{eq:edge_encoder}
    h_{ij}^{(0)} = \mathrm{LayerNorm}\left(\mathrm{MLP}\left(\mathrm{LayerNorm}\left(\sum_{k\in\{i,j\}}\mathrm{MLP}\left(h_k^{(0)}\right)\right)\right)\odot \Gamma\left(\hat{\mathbf{G}}_{ij},\mathrm{RBF}(r_{ij})\right)\right),
\end{equation}
where $\hat{\mathbf{G}}_{ij}\propto\hat{\mathbf{r}}_{ij}\otimes\hat{\mathbf{r}}^T_{ij}$ is the edge's normalized gyration tensor expressed in the traceless symmetric representation. See Section \ref{SI:convention}.

After these, all layers follow a residual update structure, $h'_r =  h_r +  g\left(h_r\right)\odot M_r(\tilde{h}_r;\{\tilde{h}_s,\tilde{h}_t\})$, where $g$ is a gating MLP with a sigmoid output that depends on the unnormed features, and $M_r(\tilde{h}_r;\{\tilde{h}_s,\tilde{h}_t\})$ is a message aggregation with $\tilde{h}_r,\tilde{h}_s,\tilde{h}_t$
being $\mathrm{LayerNorm}$ed receiver, sender, and intermediary features, respectively. In general, 
\begin{equation*}
M_r\left(\tilde{h}_r;\{\tilde{h}_s,\tilde{h}_{t}\}\right)=\sum_{s,t\in\mathcal{N}_t(r)}\mathrm{MLP}\left(\tilde{h}_r,\tilde{h}_s,\tilde{h}_{t}\right)\odot\Gamma(\mathbf{R}_r,\mathbf{R}_s),
\end{equation*}
where $(\mathbf{R}_r,\mathbf{R}_s)$ is some real-space geometric relation between receiver and sender. In node-to-node and node-to-edge messages, we use the same convention as in Eqs. \ref{eq:node_encoder} and \ref{eq:edge_encoder}, respectively. In edge-to-edge messages, $(\mathbf{R}_r,\mathbf{R}_s)\equiv\left(\hat{\mathbf{G}}_{rs}, P(\cos\left[\mathbf{G}_r,\mathbf{G}_s\right])\right)$, where $\hat{\mathbf{G}}_{rs} = (\mathbf{G}_r-\mathbf{G}_s)/\|\mathbf{G}_r-\mathbf{G}_s\|$ is the unit-norm relative gyration tensor and we project $\cos\left(\mathbf{G}_r,\mathbf{G}_s\right)=\mathrm{Tr}\left(\hat{\mathbf{G}}_r^T\hat{\mathbf{G}}_s\right)$ onto a basis of Legendre polynomials, $\{P_n(x)\}$, since the gyration tensors contain angular information.

At the end of a messaging cycle, we apply a feed-forward (FF) residual update following the expression
\begin{equation*}
    h_r' = h_r + g(h_r)\odot 
    \mathrm{MLP}\left(\tilde{h}_r, W\cdot\mathrm{EMBED}(Z_r)\right),
\end{equation*}
where we include a linear transformation of the initial embedding of the atomic species to serve as a chemical reminder (but that term is omitted for edges' FF).

Finally, the prediction of the model for the properties of a given atom, $\hat{y}_i=(\hat{N}_i,\hat{\lambda}_i,\hat{\boldsymbol{\mu}}_i,\hat{\mathbf{Q}}_i)$ is a readout that depends on an atomic head at the end of each layer:
\begin{equation*}
    \hat{y}_i^{(l)} = W\cdot\mathrm{MLP}\left( \mathrm{LayerNorm}\left(h_i^{(l)}\right) \right),\ \ 
    \text{and} \
    \hat{y}_i = \sum_l \alpha^{(l)} \hat{y}_i^{(l)},
\end{equation*}
where the layer weights $\alpha$ come from a softmax, $\alpha^{(l)} = \exp(\omega^{(l)}) / \sum_{l'}\exp(\omega^{(l')})$.

\section{Experiments}
\label{sec:experiments}

\subsection{Atomic property prediction}

Our objective was to find out which architectural features offered the best tradeoff between computational cost and quality of inferred QTA properties with respect to the ground truth distribution. In order to find the best candidate for this inference task, we trained a series of E(3)-equivariant and non-equivariant models for QTA prediction, both with fully connected and cutoff based models, ensuring all atoms had at least 2 neighbors within cutoff (see Fig. \ref{fig:rdfs_training}). We notice that the models' cutoff radii are longer than the one used for clustering the atomic environments and show that this does not incur serious data leakage in Appendix \ref{SI:cutoff_leakage}. Hyperparameter optimization was run for a first batch of experiments, after which we further tuned width and depth of the models to keep sensible, matched compute cost across models (see Appendix \ref{SI:compute}). We trained for a number of epochs between 1500 and 3000, keeping training wall-time below 2 days per fold. Nevertheless, all models plateaued after epoch 1000. The parameters with the lowest validation loss were used for each fold's predictions.

\begin{table}[h]
\caption{Architectural details of the main models in the discussion. E-models are equivariant. Data augmented models saw randomly rotated molecules every epoch.}
\label{tab:model_details}
\centering
\begin{tabular}{@{}lcccc@{}}
\toprule
        & \multicolumn{4}{c}{Architectural details}                 \\ \cmidrule(l){2-5} 
Models & Cutoff (Bohr) & Max. n.n. & Depth (layers) & Data augmented \\ \midrule
EGFC    & -             & -         & 4             & - \\
SFC2    & -             & -         & 7             & Yes \\
SGFC    & -             & -         & 7             & No \\
SG-8-12 & 8             & 12        & 7             & Yes \\
SG-8-5  & 8             & 5         & 7             & Yes \\
EGNN    & 5.25          & 5         & 4             & - \\ 
SGNN    & 5.25          & 5         & 7             & No \\
\bottomrule
\end{tabular}
\end{table}

\subsection{Molecular properties from inferred QTA properties}

We ran experiments on molecular property prediction on the AIMEl subset with a molecular version of QT-Net that used a 3.5 Bohr cutoff with 4 n.n. and 4 layers. The resulting receptive field reaches 98.3\% of the atoms within a 6-Bohr environment while deliberately limiting model capacity. Each model was trained both with (`informed') and without (`blind') ground truth QTA properties as input. Every model is fed geometry through QT-Net's geometric filters, so an informed advantage cannot come from geometry implicit in the QTA properties. Both variants were trained with identical hyperparameters for 2\,000 epochs on molecules randomly rotated at every epoch, under the 5$\times$5 CV of Sec.~\ref{sect:CV} at four training fractions (200 models in total), and we kept the parameters with the lowest validation loss.

An ensemble of 5 models with the  atomic QT-Net architecture was re-trained on the whole ground truth in AIMEl, having different validation sets for each member of the ensemble for early stopping. We use this ensemble to infer QTA properties for each of the atoms in the remainder of QM9. The quality of the multipoles and localization indices inferred by this ensemble was tested by two different experiments. In the first, we directly sum the inferred atomic contributions to the molecular dipole moments, and compare the resulting molecular dipole moments against the ground truth in QM9. In the second, every molecular model trained on AIMEl was evaluated, without retraining, on the 98\,096 molecules of the remainder. The informed variant read the QTA properties inferred by the ensemble, normalized with the statistics of its own training set.

\section{Discussion}

\subsection{Comparison of QTA property prediction on OOD atomic environments}\label{sec:QTA}

For each model, element, and QTA property, we report Lin's concordance correlation coefficient (CCC), defined as $\rho_c = 2\ \mathrm{Cov}(y,\hat{y})/(\sigma_y^2 + \sigma_{\hat{y}}^2 + (\mu_y - \mu_{\hat{y}})^2)$. This metric jointly penalizes deviations from the identity line and mean bias, making it more sensitive to always-positive quantities such as $|\boldsymbol{\mu}|$ and $|\mathbf{Q}|$.  Our discussion revolves around the interplay among cutoff, number of nearest neighbors (n.n.) and architectural constraints. The statistical validity of the comparison gets slightly compromised when digging into the peculiarities of each species, and therefore we defer this part of the discussion to Appendix \ref{SI:augment}. More architectures were tested (including topological message passing~\cite{bodnar2021weisfeiler,crespo2025topomole} and tensor product layers), which offered no advantage at increased computational cost. Interested readers are referred to the repository for more information.

\begin{table}[h]
\caption{Tukey p-value matrix for CCC values averaged over atom types and properties, comparing data augmented, non-equivariant models with equivariant models. Values above 0.05 (marked in \textbf{bold}) indicate a statistically insignificant difference in performance. The best performing models are determined based on CCC mean value and statistical significance of the difference: \colorbox{gold!50}{1st}\,\colorbox{gray!25}{2nd}\,\colorbox{orange!30}{3rd}.}
\label{tab:p-value_general}
\centering
\begin{tabular}{@{}lccccccc@{}}
\toprule
                           & \multicolumn{5}{c}{Tukey p-values}                    &                                    \\ \cmidrule(lr){2-6}
\multicolumn{1}{c}{Models} & EGFC   & SFC2    & SG-8-12 & SG-8-5 & EGNN   & CCC ± 95\% CI             \\ \midrule
EGFC                       & 1.0000 & 0.0010   & 0.0010  & \textbf{0.5729} & 0.0010 & \colorbox{gray!25}{0.798$\pm$0.006} \\
SFC2                       & 0.0010 & 1.0000   & \textbf{0.8968}  & 0.0010 & 0.0010 & \colorbox{gold!50}{0.895$\pm$0.006} \\
SG-8-12                    & 0.0010 & \textbf{0.8968}  & 1.0000  & 0.0010 & 0.0010 & \colorbox{gold!50}{0.888$\pm$0.006} \\
SG-8-5                     & \textbf{0.5729} & 0.0010  & 0.0010  & 1.0000 & 0.0010 & \colorbox{gray!25}{0.809$\pm$0.013} \\
EGNN                       & 0.0010 & 0.0010  & 0.0010  & 0.0010 & 1.0000 & \colorbox{orange!30}{0.754$\pm$0.028} \\ \bottomrule
\end{tabular}
\end{table}

The results in Table \ref{tab:p-value_general} show that non-equivariant, rotationally augmented models outperform the equivariant counterparts significantly. Looking at Table \ref{tab:model_details}, one would be tempted to attribute this advantage to the greater depth of the scalar models.  We argue that this depth is not being invested in passing chemical information, but rather in polishing the purely geometrical information that equivariant models understand by design. Interested readers are referred to Appendix Section \ref{SI:augment}, where we discuss this possibility based on results from non-augmented models. 

With the effect of the layer depth aside, the results suggest that the connectivity beyond 5 neighbors is mostly providing the non-equivariant models with geometric information, rather than chemical. On the one hand, notice that SG-8-5 performs on-par with EGFC. On the other, SG-8-12 outperforms EGFC and matches SFC2. Therefore, the increased amount of neighbors of SG-8-12 against SG-8-5 is helping the model achieve better geometric awareness, which doesn't improve beyond 12 neighbors. Given these results, we chose the SG-8-12 architecture for QT-Net.

\subsection{Atomic environments determine the accuracy of molecular predictions}\label{sec:inference}

To test the selected architecture on unseen molecules, we ran a physics-based test where we computed the molecular dipole moments for the 98\,096 molecules of the remainder from their inferred atomic contributions. Following \citet{matta2007introduction}, the molecular dipole moment can be obtained from the output of QT-Net for each atom $\Omega$ as $\boldsymbol{\mu} = \sum_\Omega \boldsymbol{\mu}(\Omega).$ 
\begin{figure}[h!]
\centering
    \includegraphics[width=0.9\linewidth]{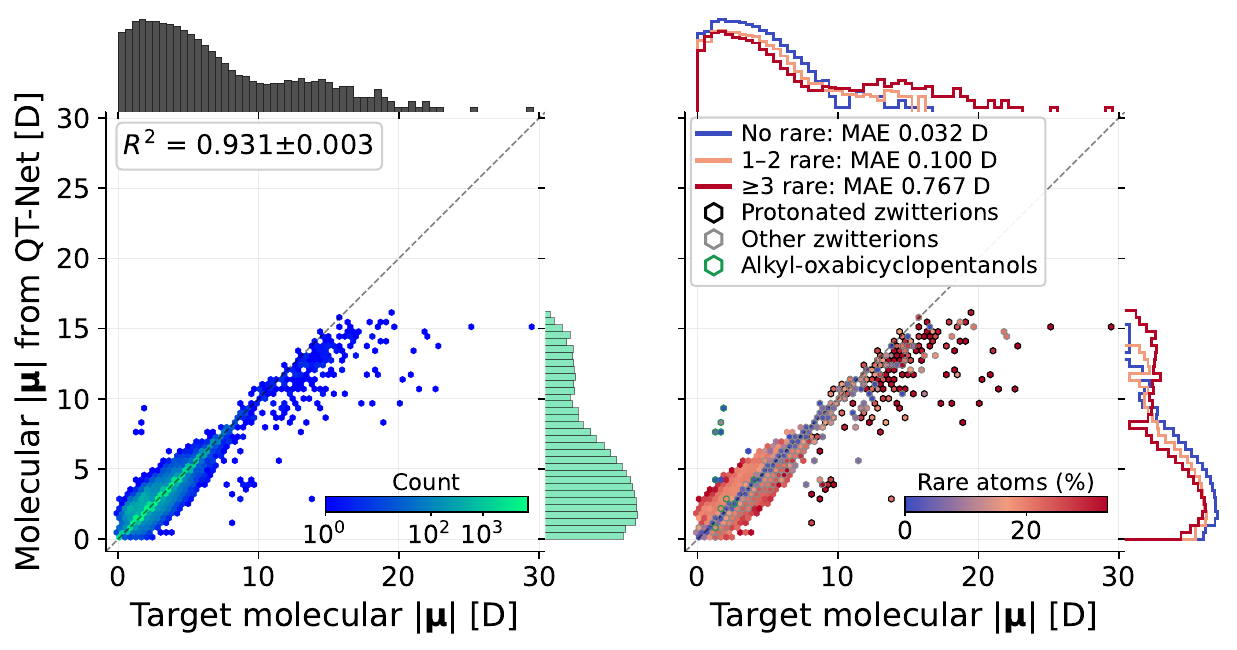}
    \caption{\textbf{(left)} Parity plot of molecular dipole moments of the 98\,096 QM9 molecules outside AIMEl, computed from the atomic contributions inferred by QT-Net. \textbf{(right)} The same plot, colored by each molecule's  percentage of rare atoms. Marginals split the molecules by their number of such atoms.}
    \label{fig:dipole_from_inferred}
\end{figure}

Although QT-Net \textit{was never trained on molecular dipoles}, the sums recover the magnitudes reported in QM9 with $R^2 = 0.931$ (Fig.~\ref{fig:dipole_from_inferred}, left). This recovery is a strong physical consistency check: because molecular dipole moments arise from the vector sum of per-atom contributions, accurate reconstruction requires that QT-Net has correctly learned not only the magnitudes but also the orientations of the inferred atomic dipoles relative to one another. This indicates that the model has captured the underlying QTAIM partitioning of the electron density and not simply memorized artifacts of the AIMEl distribution.

Regarding molecular property prediction, our experiments show that QTA descriptors do not provide significant advantages in general. We dig deeper into this analysis in Appendix \ref{SI:descriptors}. Our analysis revealed that the atomic environments seen during training matter more than the descriptors. As each molecule is predicted by models that saw its atomic environments or chemotype and by models that did not (Sec.~\ref{sect:CV}), the cost of each holdout can be measured on the same molecules.

\begin{figure}[h!]
    \centering
    \includegraphics[width=\linewidth]{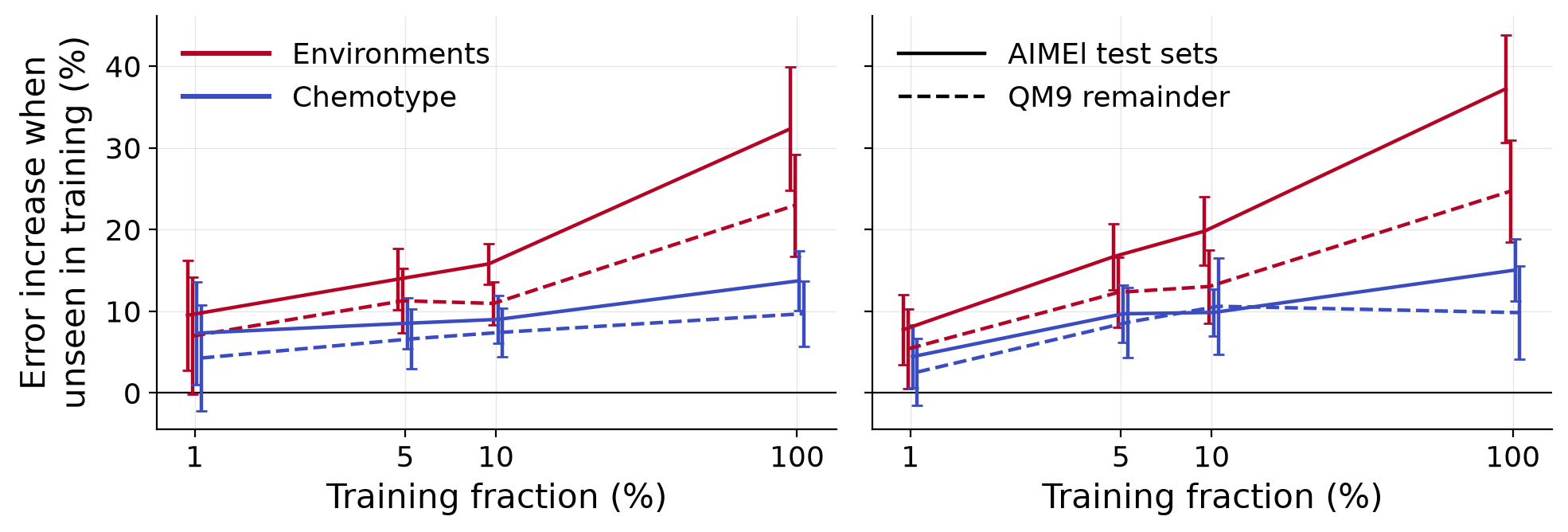}
    \caption{Error increase on the same molecules when their atomic environments or their chemotype were unseen in training, relative to the models that saw them, averaged over the four targets (Appendix \ref{SI:descriptors}). \textbf{(left)} Blind and \textbf{(right)} informed models. Fold means with 95\% $t$-intervals.}
    \label{fig:same_molecule}
\end{figure}

In both studies, we found the distance to the training distribution in atomic -- rather than molecular -- chemical space to be the main source of error. We measure the distance of an atom to a training set as the distance to its 15th-nearest training atom of the same element in the 6-Bohr descriptor space (see Appendix ~\ref{SI:QM9_clustering}), and say that an atom is `far from AIMEl' or `rare' when it sits beyond the 99th percentile of the same distance for in-distribution atoms. For the reconstructed dipoles, the errors concentrate where the remainder leaves the atomic chemical space of AIMEl. Molecules with 0, 1--2 and $\geq$3 atoms far from AIMEl have MAEs of 0.032, 0.100 and 0.767\,D (Fig.~\ref{fig:dipole_from_inferred}, right). For 855 of the molecules with $\geq$3 rare atoms, the reconstruction is nearly exact. The rest of the molecules in this group are alkynes (14 molecules in AIMEl against 17\,468 in the remainder 77\% of the error) and zwitterions. QT-Net thus transfers to rare environments of familiar chemistry, but not to chemistry its training set lacks (see Appendix~\ref{SI:dipole_errors}). For the molecular models, molecules with $\geq$3 atoms far from their training set have 1.3 to 1.9 times the error of molecules with none on the AIMEl test sets, and up to 4.2 times on the remainder (see Fig. \ref{fig:error_vs_novelty}).

\section{Limitations}\label{sec:Limitations}

Here, we justify the usage of a common held-out set for evaluating the 25 folds of the models in Section \ref{sec:QTA}. In standard molecular benchmarks, scaffolds are randomly assigned to folds, so the expected test-set difficulty is approximately equal across folds. Between-fold variance therefore reflects sampling noise, and the ANOVA exchangeability assumption holds approximately. At the atomic level, atomic environments play the role of ``scaffolds'', but they co-occur within molecules: specific cluster labels are mutually enriched, so the presence of one label in a molecule raises the probability of finding another. 

During training, showing the model an atomic label that is highly enriched with a held-out label would expose the model to atomic environments that co-reside in the same molecules as the held-out environments, allowing it to learn their structural context indirectly. Mutually enriched label sets must therefore be withheld together as a unit. In the AIMEl data regime, this constraint severely limits the number of admissible holdout combinations and precludes randomizing label assignments across repeats. 

As a result, the Tukey HSD comparisons presented in this work are conditional on the chosen holdout and cannot be interpreted as claims of general OOD superiority of either design, nor of general superiority of non-equivariant over E(3)-equivariant models. We use them as guidance for deciding on the final QT-Net design for the inference and downstream tasks.

\section{Conclusion}

Our work shows a statistically thorough analysis of the performance difference between E(3)-equivariant and purely scalar, rotationally augmented models for the task of learning scalar and tensorial QTA properties. Our results suggest that scalar models benefit from densely connected graphs for capturing geometrical information, especially when data augmentation is used. We therefore justify a non-equivariant design for the introduced architecture, QT-Net, which we demonstrate can reconstruct molecular dipole moments outside the set of training molecules from inferred atomic contributions, with errors concentrated in environments that AIMEl barely samples, most of all alkynes and zwitterions.

We want to emphasize that OOD evaluation of AI models for QTA properties requires careful labeling and tracking of the atomic environments. The properties of an atom depend on its environment, and the same element can appear in chemically distinct environments within a molecule. Our approach shows that SOAP-based clustering can separate these chemically different environments, and enables us to construct OOD splits by withholding molecules that contain environments absent from the training set. Furthermore, using metrics that specifically measure model performance on these unseen environments enables us to design an architecture based on faithful metrics of OOD performance. Additional results in Section \ref{SI:test} (specifically in Fig. \ref{fig:Tukey_test_CCC}) show that \textbf{metrics that account for all atoms in a set of test molecules lead to overconfident results that do not evaluate a model's true OOD performance.} This pitfall has been repeatedly observed in the AI-for-QCT literature (see Section \ref{sec:related}). 

At the molecular level, the coverage of atomic environments matters more than atomic descriptors. Holding out atomic environments and chemotypes at the same time, from 803 training molecules on we found that unseen environments raise the error of the same molecule by 11--37\%, 1.2--2.5 times the cost of an unseen chemotype. QTA properties, whether ground truth or inferred, do not make up for them. The atomic environments seen in training are therefore a more important hinge of molecular property prediction than atomic descriptors, and evaluations of molecular models should track them as carefully as evaluations of atomic ones.

Our results suggest that further progress on QTA property prediction is now primarily a data problem rather than a representation or architectural challenge: across the four properties studied here, the same architecture achieves CCC values ranging from $\sim$0.4 to $\sim$0.99 depending on element and property, indicating that the bottleneck has shifted from how to model these targets to which targets to model and on what chemistry. The dipole analysis makes this concrete: alkynes, with 14 samples in AIMEl against 17\,468 in the remainder, account for 77\% of the error of the reconstructed dipoles. Extending the evaluation protocols introduced here to other quantum-mechanically derived atomic descriptors—such as conceptual DFT reactivity indices, natural bond orbital populations, IQA decompositions—and to broader regions of chemical space is, in our view, the natural next step.

\section{Code and data availability}\label{Sec:code}
Code and data for this project are available at \url{https://github.com/pablomcrespo/QT-Net}. It includes all seeds and information to reproduce the experiments of this work, besides relevant code to reproduce the statistical analysis. The source data is referenced in the text, but we include compressed versions of our curated files (mainly for reproducibility with the same cluster labels). The QTAIM properties inferred by QT-Net are released in the repository, too.

\begin{ack}
PMC and RB acknowledge funding from the Intel-Merck joint university research center for AI-Aware Pathways to Sustainable Semiconductor Process and Manufacturing Technologies (AWASES). RM acknowledges funding provided by the Wallenberg AI, Autonomous Systems, and Software Program (WASP), supported by the Knut and Alice Wallenberg Foundation. PMC acknowledges Lynne Ressel for the inspiration in naming QT-Net. The computations and data storage were enabled by resources provided by Chalmers e-Commons and by the National Academic Infrastructure for Supercomputing in Sweden (NAISS), partially funded by the Swedish Research Council through grant agreement no. 2022-06725. The authors declare no competing interests.
\end{ack}

\section*{Author contributions}
PMC performed the main research activities, collected and curated the dataset, analyzed the results, prepared the visualizations, and wrote the original draft of the manuscript. SR and RB contributed to the planning of the cross validation experiments, statistical analysis and visualizations. MG, SM, RSJ and VKN contributed with methodological discussion and supervision. RM managed resources and acquired funding for the project with support from SM and VKN. MR and RM contributed through conceptual input, critical discussion, supervision, and manuscript review and editing.
\newpage

{
\small
\bibliographystyle{plainnat}
\bibliography{references}
}


\appendix

\newpage

\section{Technical appendices and supplementary material}

\subsection{Training and computational details}\label{SI:compute}

We report training and compute statistics for all models compared in this work in Table~\ref{tab:training_details}. We characterize per-batch compute using GFLOPs (giga floating-point operations, $10^9$ FLOPs), measured as the cost of a single forward pass through the model on one training batch. EGFC model trained on A100fat GPU for about 2 days per fold. All other models trained on either A40 or A100 (based on GPU availability) and took from 12 to 36 hours per fold.

\begin{table}[h!]
\centering
\caption{Training and computational details for the models compared in Section~\ref{sec:experiments}. Parameter counts were chosen to keep model sizes approximately matched across architectures, while batch size and epoch count were tuned per model based on convergence behavior. Abbreviations: learning rate (LR); weight decay for AdamW (WD).}
\label{tab:training_details}
\begin{tabular}{lcccccc}
\toprule
\multicolumn{1}{c}{Models} & \# Params. & \multicolumn{1}{l}{Batch Size} & GFLOPs/Batch & Epochs & LR & WD \\ \midrule
\textbf{EGFC}              & 486\,786   & 128                            & 17.88        & 1500   & 1.6e-3 & 1.2e-4\\
\textbf{SFC2}              & 686\,081   & 256                            & 20.30        & 2000   & 2.7e-3 & 1.6e-4\\
\textbf{SGFC}              & 487\,598   & 128                            & 8.67         & 3000   & 2.7e-3 & 1.6e-4\\
\textbf{SG-8-12}           & 686\,049   & 256                            & 9.25         & 2000   & 3.8e-3 & 1.1e-4\\
\textbf{SG-8-5}            & 686\,049   & 2048                           & 17.99        & 2000   & 3.8e-3 & 1.1e-4\\
\textbf{EGNN}              & 615\,090   & 512                            & 8.89         & 3000   & 3.8e-3 & 1.1e-4\\
\textbf{SGNN}              & 677\,089   & 1024                           & 8.96         & 3000   & 5.5e-3 & 9.1e-4\\ \bottomrule
\end{tabular}
\end{table}

EGFC, EGNN, SGFC, and SGNN ran in a first round of experiments that used for loss function
\begin{equation*}
    \mathcal{L}=\frac{1}{N_{at}}\sum_{i=1}^{N_{at}}\frac{1}{4}\left[(\hat{N}_i-N_i)^2+(\hat{\lambda}_i-\lambda_i)^2+
    \frac{\sum_{\alpha=1}^3(\hat{\mu}_i^{(\alpha)}-\mu_i^{(\alpha)})^2}{3}+
    \frac{\sum_{\alpha=1}^5(\hat{Q}_i^{(\alpha)}-Q_i^{(\alpha)})^2}{5}
    \right],
\end{equation*}
where $N_{at}$ is the total number of atoms in the training fold. For the rotationally augmented models, the training was slightly refined, adding per-element weights $w_{Z_i}$ and modifying the quadrupole term to account for the Frobenius weights:
\begin{equation*}
    \mathcal{L}=\frac{1}{\sum_{i=1}^{N_{at}}w_{Z_i}}\sum_{i=1}^{N_{at}}\frac{w_{Z_i}}{4}\left[(\hat{N}_i-N_i)^2+(\hat{\lambda}_i-\lambda_i)^2+
    |\hat{\boldsymbol{\mu}}_i-\boldsymbol{\mu}_i|^2+
    \|\hat{\mathbf{Q}}_i-\mathbf{Q}_i\|_F^2
    \right].
\end{equation*}
The weight of each atom was normalized so that the sum of all atoms of a same element in the fold summed to 1, following $w_{Z_i}=\sqrt{N_{at}/N_{Z_i}}$. 

For all models, the target quantities were z-score normalized, using mean and standard deviation for scalar properties and dividing by root mean square for tensor properties. 

All points in the comparison stand, given the direct comparison between non-rotationally augmented and equivariant models in Section \ref{SI:augment}.

\subsection{Conventions}\label{SI:convention}

Our convention for representing a traceless tensor in Cartesian coordinates is

\begin{equation*}
    [T_{xx},T_{xy},T_{xz},T_{yy},T_{yz},T_{zz}] \rightarrow \ [T_{xy},T_{xz},T_{yz},T_{an},T_{zz}],
\end{equation*}
where the anisotropic component is obtained from $T_{an}=\frac{T_{xx}-T_{yy}}{2}$. The equivariance-preserving Frobenius norm is therefore computed as 
\begin{equation}\label{eq:frob_norm}
    \|T\| = \sqrt{2(T_{xy}^2+T_{xz}^2+T_{yz}^2+T^2_{an})+\frac{3}{2}T_{zz}^2}.
\end{equation}

\subsection{Clustering of AIMEl atomic environments}

\begin{figure}[h!]
    \centering
    \includegraphics[width=1.0\linewidth]{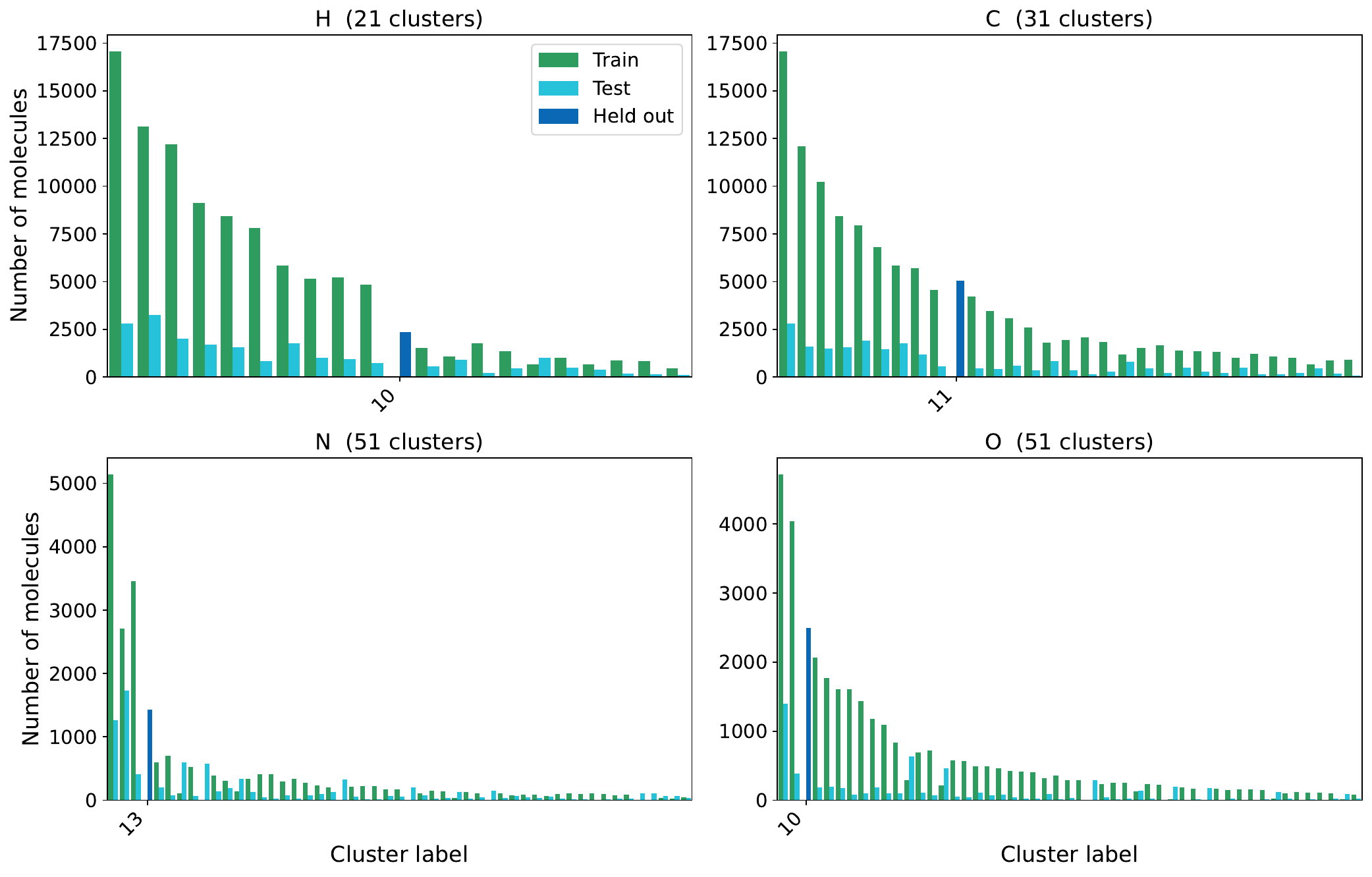}
    \caption{Number of molecules where an atomic environment occurs in the whole AIMEl subset. Only held out environment labels are included for clarity.}
    \label{fig:atom_label_spread}
\end{figure}

Element-specific HDBSCAN clustering parameters and resulting cluster counts 
are reported in Table~\ref{tab:clustering}, and the distributions of cluster 
labels across molecules in the dataset is shown in 
Figure~\ref{fig:atom_label_spread}.

\begin{table}[h]
\caption{Element-specific details of HDBSCAN clustering. In all cases, we used the excess of mass method and no cluster merging.}
  \label{tab:clustering}
  \centering
\begin{tabular}{@{}lcccc@{}}
\cmidrule(l){2-5}
\multicolumn{1}{c}{}   & \textbf{H} & \textbf{C} & \textbf{N} & \textbf{O} \\ \midrule
Number of atoms        & 278\,943    & 199\,660    & 30\,805     & 40\,894     \\
PCA components         & 20         & 35         & 30         & 20         \\
Variance retained (\%) & 99.11      & 99.01      & 99.18      & 99.04      \\
Min. cluster size      & 850        & 850        & 100        & 100        \\
Min. samples           & 200        & 200        & 30         & 30         \\
Number of clusters     & 20         & 30         & 50         & 50         \\
Noise points (\%)      & 8.2        & 5.95       & 16.69      & 16.5       \\ \bottomrule
\end{tabular}
\end{table}

\begin{figure}[h!]
    \centering
    \includegraphics[width=1.0\linewidth]{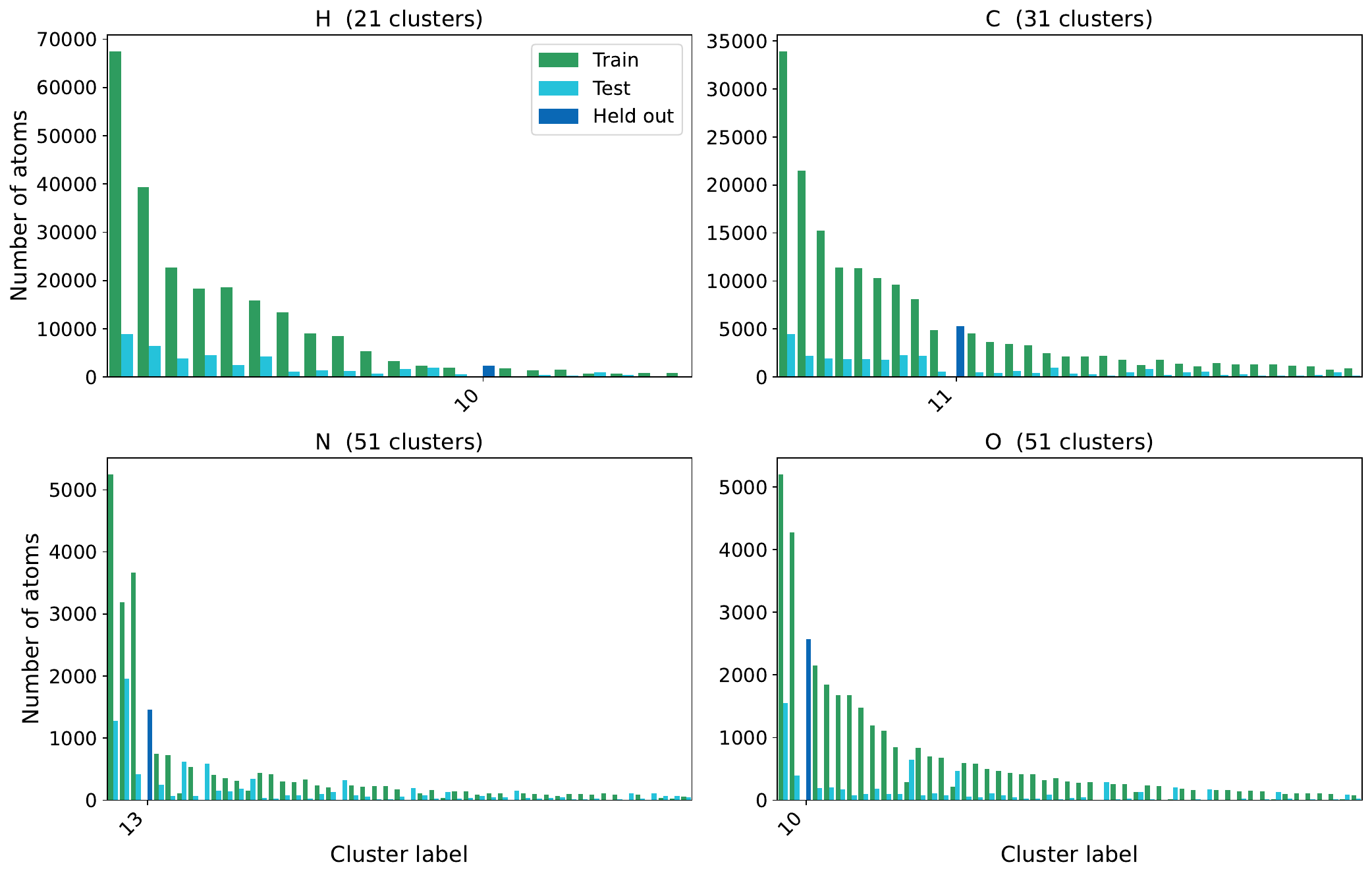}
    \caption{Number of atoms in each atomic environment. Only held out environment labels are included for clarity.}
    \label{fig:atom_label_count}
\end{figure}

\clearpage

\subsection{Clustering of QM9 atomic environments and chemotypes}\label{SI:QM9_clustering}

\begin{figure}[h!]
    \centering
    \includegraphics[width=0.95\linewidth]{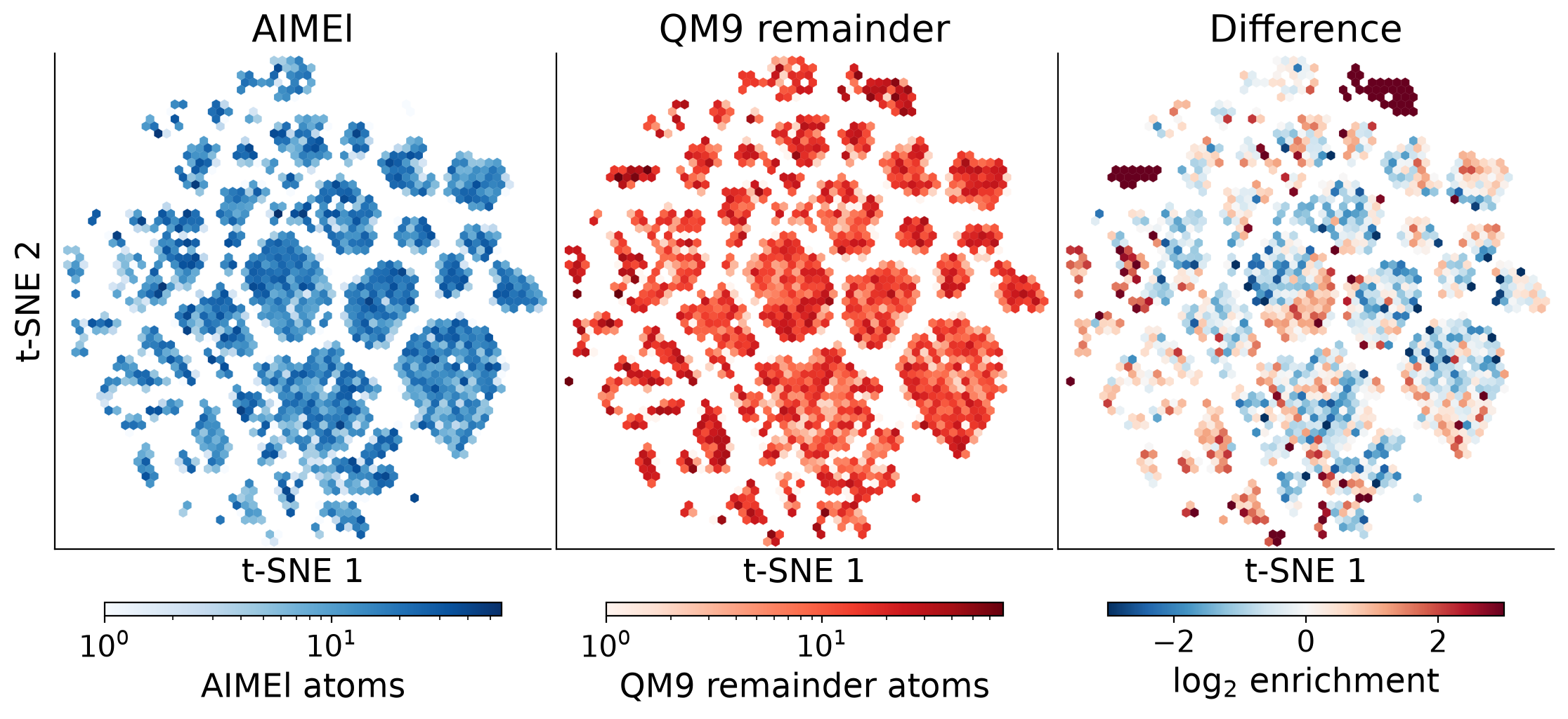}
    \caption{Atomic environments of C: t-SNE (perplexity 50, PCA initialization) of 20\,000 AIMEl and 20\,000 remainder atoms drawn at random, in the 6-Bohr SOAP space. From left to right: hexagonal-bin counts of the AIMEl atoms, the QM9 remainder atoms, and the log$_2$ enrichment of the remainder over AIMEl in each bin, i.e.\ the log$_2$ ratio of the remainder's share of its sample to AIMEl's share (red: denser in the remainder; blue: denser in AIMEl; bins with fewer than three atoms left blank). t-SNE preserves neighborhoods, not distances or densities across the map.}
    \label{fig:tsne_C}
\end{figure}

\begin{figure}[h!]
    \centering
    \includegraphics[width=0.95\linewidth]{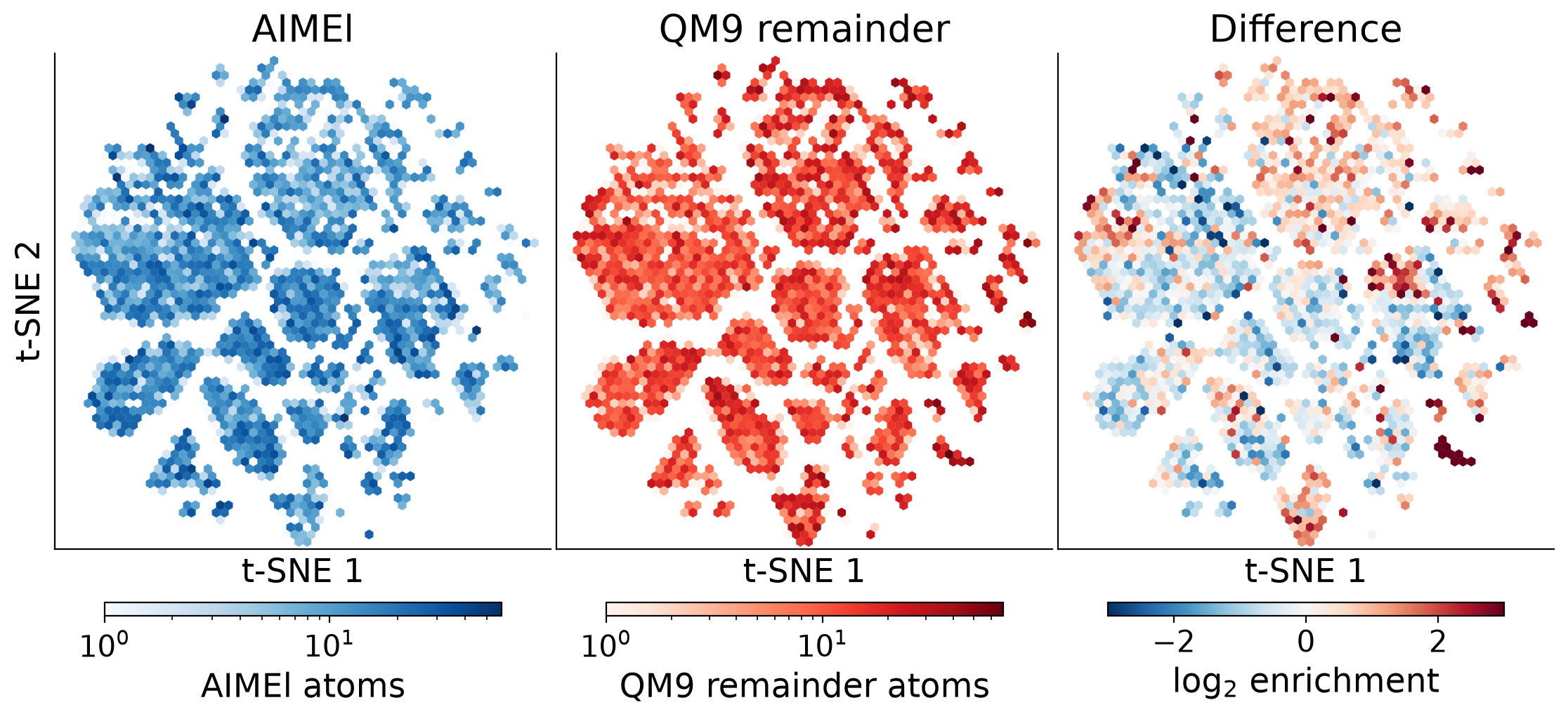}
    \caption{Atomic environments of H, as in Fig.~\ref{fig:tsne_C}.}
    \label{fig:tsne_H}
\end{figure}

\begin{figure}[h!]
    \centering
    \includegraphics[width=0.95\linewidth]{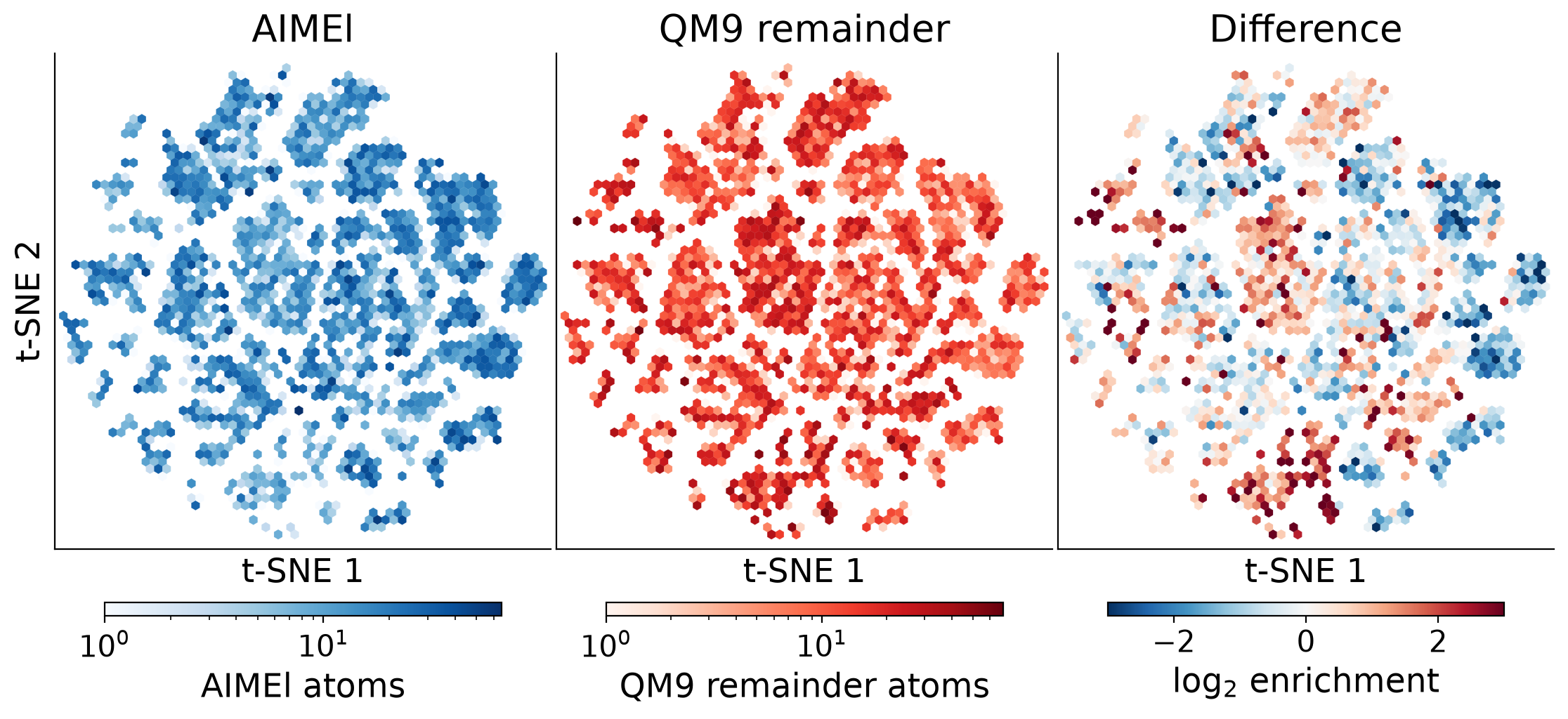}
    \caption{Atomic environments of O, as in Fig.~\ref{fig:tsne_C}.}
    \label{fig:tsne_O}
\end{figure}

\begin{figure}[h!]
    \centering
    \includegraphics[width=0.95\linewidth]{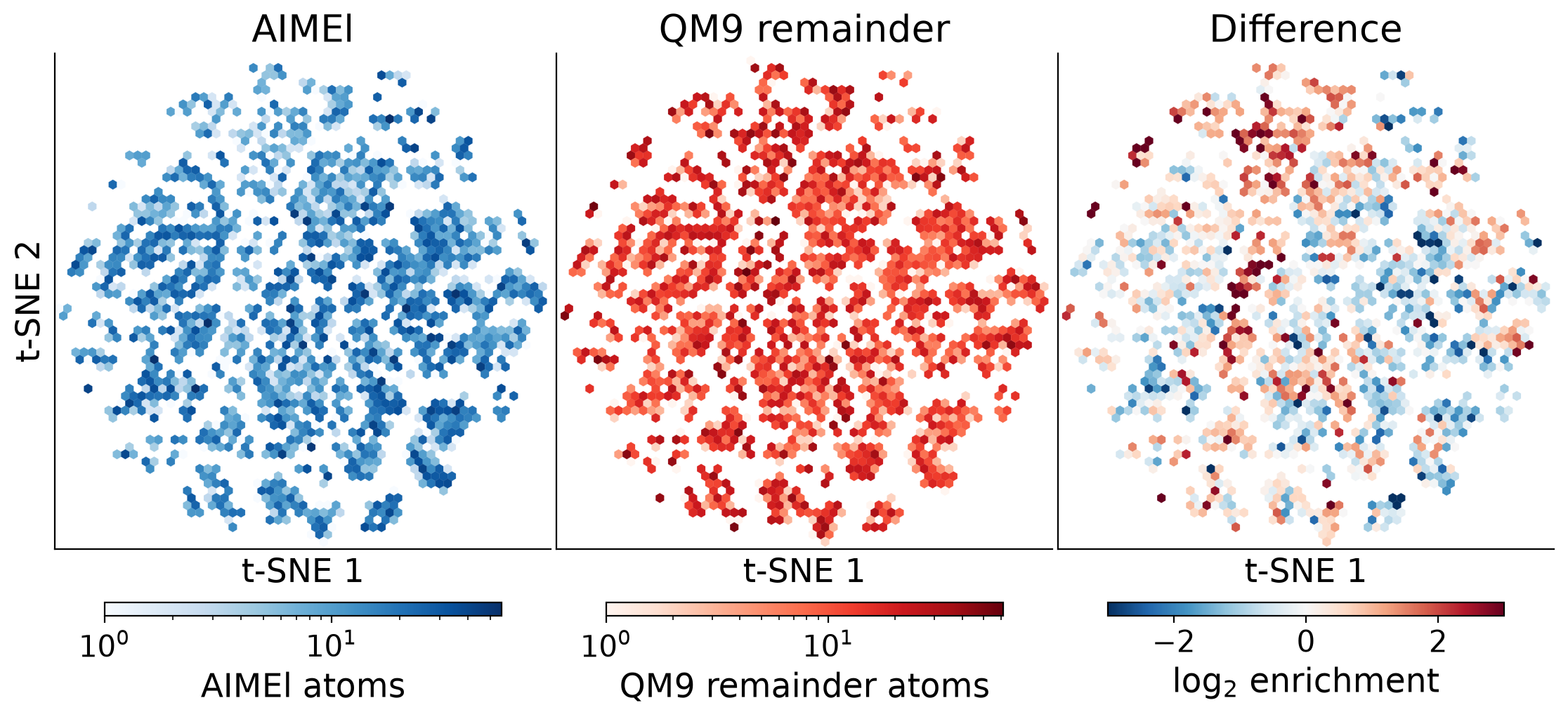}
    \caption{Atomic environments of N, as in Fig.~\ref{fig:tsne_C}.}
    \label{fig:tsne_N}
\end{figure}

\textbf{Atomic environments.} Element-specific details for the clustering over all atoms in the 128\,908 molecules (30\,812 in AIMEl and 98\,096 in the remainder; 2\,333\,341 atoms) of the working dataset for molecular experiments are reported in Table~\ref{tab:clustering_qm9}. Because AIMEl and the remainder were clustered together, they share one set of labels. Seven environments (three of C, one of N and three of O) have no atom in AIMEl. These labels are numbered independently of the 4.5-Bohr labels of the atomic holdout; a shared name such as O\_10 is a coincidence. Figures~\ref{fig:tsne_C}--\ref{fig:tsne_N} map the environments of each element.

\textbf{Distance to AIMEl.} We measure how far an atomic environment lies from a training set in the PCA projection of the 6-Bohr SOAP vectors. The distance of an atom to a training set is its Euclidean distance to the 15th-nearest training atom of the same element. We calibrate it on in-distribution atoms of molecules outside the reference, and say that an atom is `far from AIMEl' or `rare' when its distance exceeds their 99th percentile. For QT-Net, the training set is AIMEl, and every AIMEl atom is scored without the atoms of its own molecule, so that 1\% of the AIMEl atoms are rare by construction. For the molecular models, the training set is each model's training molecules and the calibration its validation molecules. Measured this way, 3.0, 7.3, 1.7 and 2.7\% of the remainder's H, C, N and O atoms are far from AIMEl, and 35.8\% of the remainder molecules contain at least one rare atom (14.9\% three or more), against 14.7\% (0.5\%) of the AIMEl molecules by chance. Every remainder molecule in the two chemotypes absent from AIMEl, and in the three that AIMEl holds once or twice, has a rare atom. 

\begin{table}[h!]
\caption{Element-specific details of the HDBSCAN clustering of the 6-Bohr 
environments of all molecules of the working dataset. In all cases, we used the 
excess of mass method and no cluster merging.}
\label{tab:clustering_qm9}
\centering
\begin{tabular}{@{}lcccc@{}}
\toprule
 & \textbf{H} & \textbf{C} & \textbf{N} & \textbf{O} \\ \midrule
Atoms in AIMEl & 278\,943 & 199\,660 & 30\,805 & 40\,894 \\
Atoms in the remainder & 920\,627 & 623\,154 & 98\,489 & 140\,769 \\
PCA components & 48 & 76 & 67 & 52 \\
Variance retained (\%) & 99.01 & 99.01 & 99.01 & 99.02 \\
Min. cluster size & 2\,500 & 850 & 300 & 250 \\
Min. samples & 100 & 50 & 5 & 10 \\
Number of clusters & 38 & 85 & 91 & 103 \\
Clusters without AIMEl atoms & 0 & 3 & 1 & 3 \\
Noise points (\%) & 22.9 & 14.1 & 20.8 & 19.7 \\ \bottomrule
\end{tabular}
\end{table}

\clearpage

\clearpage

\textbf{Environment folds.} Each environment fold is made of four labels, one per element, that co-occur in the same molecules. Following the logic of the atomic holdout of Sec.~\ref{sec:dataset}, withholding a fold removes from training every molecule that carries one of its labels so that environments that co-occur with the withheld ones cannot leak into training. Each fold was grown around a seed environment of N or O, the scarce elements, then adding C and H environments whose atoms co-occur with the seed in at least 95\% of the seed atoms' molecules, and then the environment of the other scarce element with the largest share of its molecules inside the resulting set. Seeds were taken rarest first, alternating between N and O, from the environments holding 1--15\% of their element's AIMEl atoms. A candidate was rejected when no C or H environment reached 95\% co-occurrence, when it would withhold more than 25\% of AIMEl, when it reused an environment of an accepted fold, when one of its environments had fewer than 50 atoms, or when more than half of its molecules already belonged to accepted folds. Noise atoms carry no label and were ignored in this design. Table~\ref{tab:env_folds} lists the five folds. They were designed on AIMEl, so none of the seven environments found only in the remainder belongs to a fold.

\begin{table}[h!]
\caption{The five environment folds. Molecules: AIMEl and remainder molecules that carry at least one of the fold's environments. Atoms: AIMEl atoms in the fold's environment of each element.}
\label{tab:env_folds}
\centering\small
\begin{tabular}{@{}clrrrrrr@{}}
\toprule
 & & \multicolumn{2}{c}{Molecules} & \multicolumn{4}{c}{AIMEl atoms} \\ \cmidrule(lr){3-4}\cmidrule(l){5-8}
Fold & Environments (seed in bold) & AIMEl & Remainder & H & C & N & O \\ \midrule
0 & H\_29, C\_52, \textbf{N\_65}, O\_10 & 2\,653 (8.6\%) & 8\,620 & 3\,211 & 3\,472 & 252 & 78 \\
1 & H\_21, C\_72, N\_26, \textbf{O\_51} & 3\,497 (11.3\%) & 12\,156 & 3\,784 & 485 & 67 & 359 \\
2 & H\_13, C\_34, \textbf{N\_27}, O\_36 & 2\,664 (8.6\%) & 9\,845 & 1\,793 & 2\,122 & 503 & 65 \\
3 & H\_28, C\_84, \textbf{N\_81}, O\_16 & 6\,700 (21.7\%) & 15\,263 & 9\,497 & 9\,546 & 497 & 85 \\
4 & H\_12, C\_79, \textbf{N\_24}, O\_19 & 5\,065 (16.4\%) & 15\,683 & 792 & 5\,176 & 584 & 202 \\ \bottomrule
\end{tabular}
\end{table}

\begin{figure}[h!]
    \centering
    \includegraphics[width=0.95\linewidth]{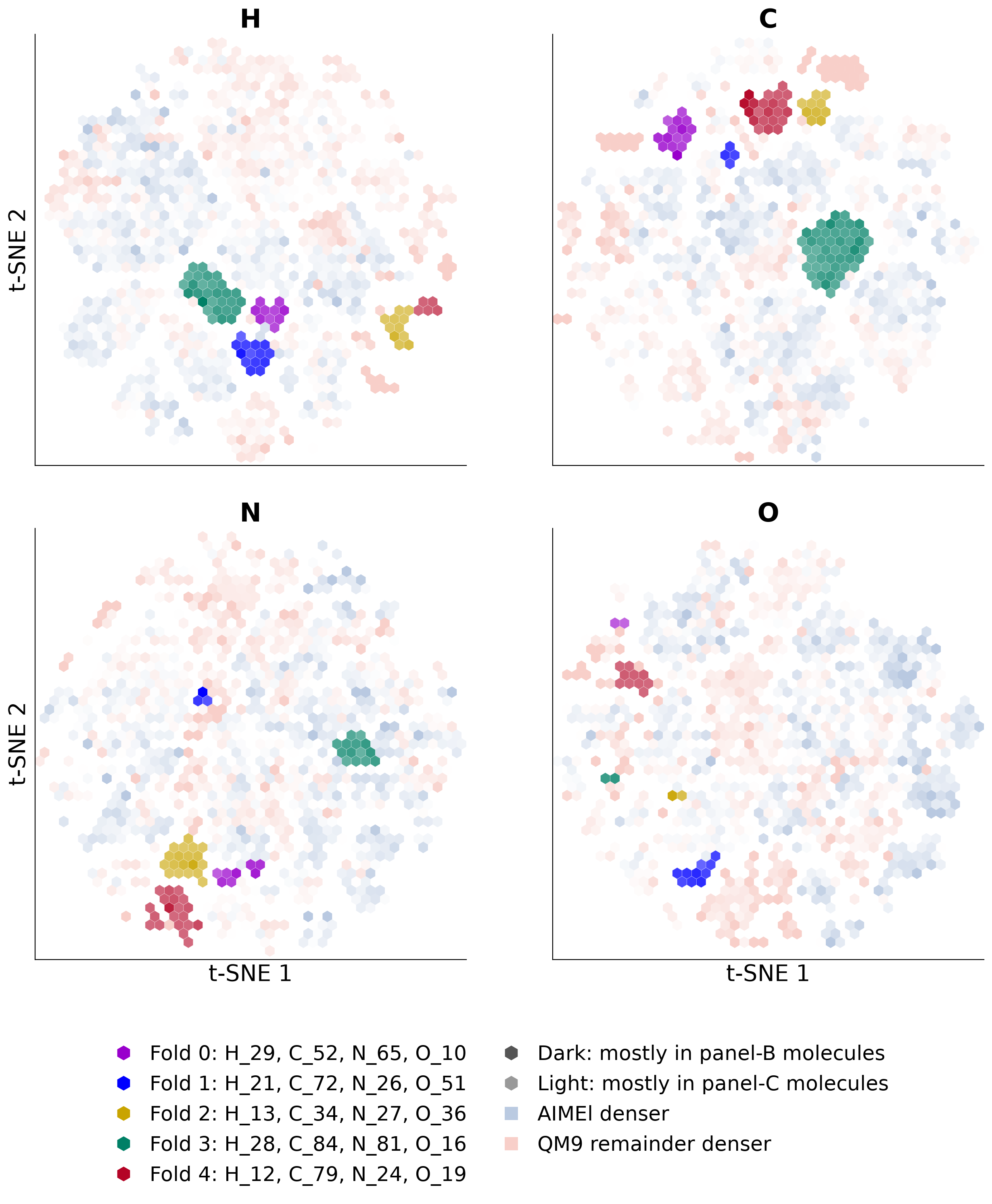}
    \caption{Environments withheld by the five folds of the first repeat on the t-SNE of each element (Figs.~\ref{fig:tsne_C}--\ref{fig:tsne_N}). Each fold withholds one environment per element. Hexagonal bins of the withheld atoms are colored by fold: dark where the atoms are mostly in panel-B molecules (chemotype also withheld), light where mostly in panel-C molecules (chemotype seen). }
    \label{fig:tsne_environments_folds}
\end{figure}

\begin{figure}[h!]
    \centering
    \includegraphics[width=0.95\linewidth]{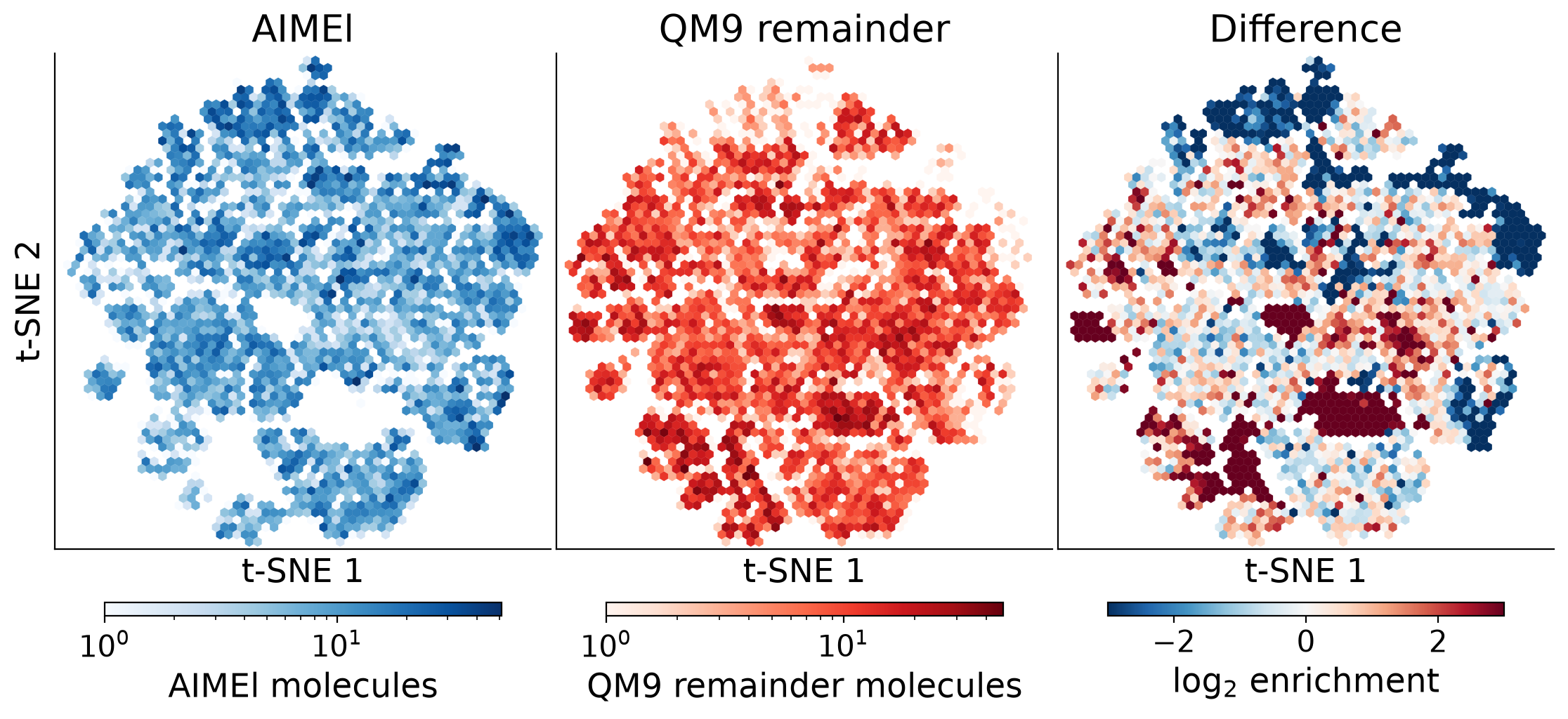}
    \caption{Chemotype space. t-SNE of 20\,000 AIMEl and 20\,000 remainder molecules drawn at random, in the space of the chemotype clustering (Morgan fingerprints, L2-normalized, 50 principal components). Panels as in Fig.~\ref{fig:tsne_C}, for molecules.}
    \label{fig:tsne_chemotypes}
\end{figure}

\begin{figure}[h!]
    \centering
    \includegraphics[width=0.75\linewidth]{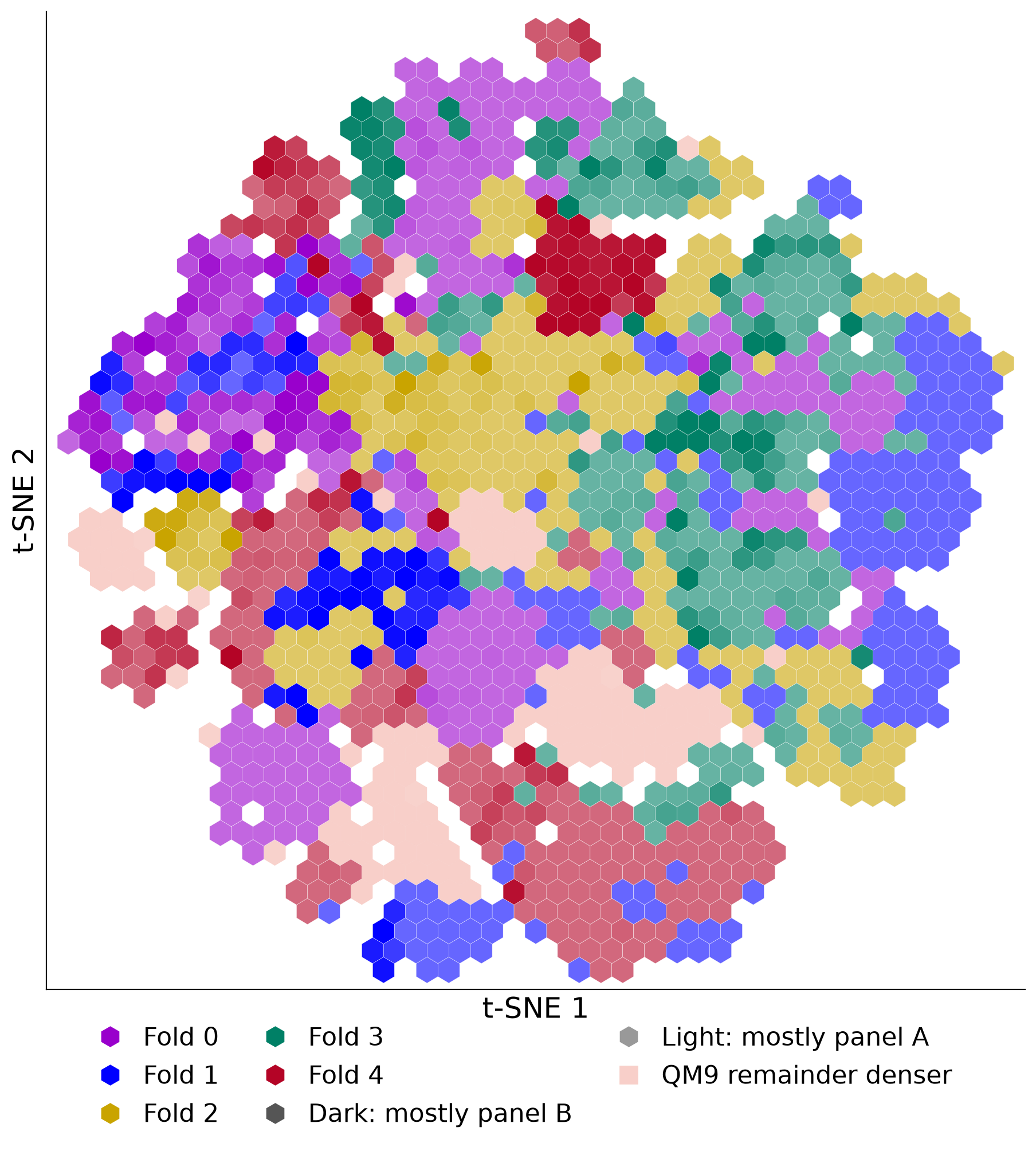}
    \caption{Chemotype groups withheld by the five folds of the first repeat, on the chemotype t-SNE (Fig.~\ref{fig:tsne_chemotypes}). Each hexagonal bin of withheld molecules is colored by the fold with most molecules in it: dark where they mostly carry the fold's withheld environments (panel B), light where they do not (panel A). 
    }
    \label{fig:tsne_chemotypes_folds}
\end{figure}

\textbf{Cross-validation.} The chemotypes of Sec.~\ref{sec:dataset} were fitted once on AIMEl and the remainder together and hold 1\,251--5\,164 molecules. Two of them contain 3\,768 terminal alkynes, which have no molecule in AIMEl. Figure~\ref{fig:tsne_chemotypes} maps the chemotype space. The 48 chemotypes present in AIMEl were dealt into five groups of whole chemotypes with balanced molecule counts (largest chemotype first, into the currently smallest group; 6\,101--6\,288 molecules per group). Fold $(r,c)$ withholds chemotype group $c$ and environment fold $(r+c) \bmod 5$, so that over the five repeats each chemotype group meets each environment fold once (Table~\ref{tab:molecular_folds}, Figs.~\ref{fig:tsne_environments_folds} and~\ref{fig:tsne_chemotypes_folds}). The test set of each fold has four panels: (A) molecules of the withheld chemotypes without withheld environments; (B) molecules of the withheld chemotypes with withheld environments; (C) molecules of other chemotypes with withheld environments; and (D) 924 molecules (3\% of AIMEl) drawn at random from the rest as an in-distribution reference. The remaining molecules form the training pool, cut to 17\,855 molecules in every fold (the size of the smallest pool) with a chemotype-stratified subsample. A chemotype-stratified 10\% of the pool served for validation and early stopping. The training fractions are chemotype-stratified subsamples of the rest, so that every chemotype labeled as seen is in training at every fraction. Test and validation sets are the same at every fraction and for both variants. On the QM9 remainder, each molecule falls into a panel depending on the training set of the model evaluating it.

\begin{table}[h!]
\caption{The 25 folds of the molecular cross-validation: the withheld chemotype group and environment fold (identified by its seed environment, Table~\ref{tab:env_folds}), and the size of each test panel. Every fold has a training pool of 17\,855 molecules and a validation set of 1\,785.}
\label{tab:molecular_folds}
\centering\small
\begin{tabular}{@{}rrrlrrrrrr@{}}
\toprule
Fold & Repeat & \shortstack{Chemotype\\group} & \shortstack{Environment\\fold (seed)} & \shortstack{Withheld\\chemotypes} & A & B & C & D & Test \\ \midrule
0 & 0 & 0 & 0 (N\_65) & 9 & 5\,256 & 1\,032 & 1\,621 & 924 & 8\,833 \\
1 & 0 & 1 & 1 (O\_51) & 8 & 4\,870 & 1\,287 & 2\,210 & 924 & 9\,291 \\
2 & 0 & 2 & 2 (N\_27) & 9 & 5\,590 & 516 & 2\,148 & 924 & 9\,178 \\
3 & 0 & 3 & 3 (N\_81) & 12 & 4\,306 & 1\,795 & 4\,905 & 924 & 11\,930 \\
4 & 0 & 4 & 4 (N\_24) & 10 & 4\,443 & 1\,717 & 3\,348 & 924 & 10\,432 \\ \addlinespace[2pt]
5 & 1 & 0 & 1 (O\_51) & 9 & 5\,957 & 331 & 3\,166 & 924 & 10\,378 \\
6 & 1 & 1 & 2 (N\_27) & 8 & 5\,927 & 230 & 2\,434 & 924 & 9\,515 \\
7 & 1 & 2 & 3 (N\_81) & 9 & 4\,876 & 1\,230 & 5\,470 & 924 & 12\,500 \\
8 & 1 & 3 & 4 (N\_24) & 12 & 5\,616 & 485 & 4\,580 & 924 & 11\,605 \\
9 & 1 & 4 & 0 (N\_65) & 10 & 5\,790 & 370 & 2\,283 & 924 & 9\,367 \\ \addlinespace[2pt]
10 & 2 & 0 & 2 (N\_27) & 9 & 5\,136 & 1\,152 & 1\,512 & 924 & 8\,724 \\
11 & 2 & 1 & 3 (N\_81) & 8 & 5\,192 & 965 & 5\,735 & 924 & 12\,816 \\
12 & 2 & 2 & 4 (N\_24) & 9 & 5\,312 & 794 & 4\,271 & 924 & 11\,301 \\
13 & 2 & 3 & 0 (N\_65) & 12 & 6\,014 & 87 & 2\,566 & 924 & 9\,591 \\
14 & 2 & 4 & 1 (O\_51) & 10 & 5\,581 & 579 & 2\,918 & 924 & 10\,002 \\ \addlinespace[2pt]
15 & 3 & 0 & 3 (N\_81) & 9 & 5\,333 & 955 & 5\,745 & 924 & 12\,957 \\
16 & 3 & 1 & 4 (N\_24) & 8 & 5\,743 & 414 & 4\,651 & 924 & 11\,732 \\
17 & 3 & 2 & 0 (N\_65) & 9 & 5\,332 & 774 & 1\,879 & 924 & 8\,909 \\
18 & 3 & 3 & 1 (O\_51) & 12 & 6\,081 & 20 & 3\,477 & 924 & 10\,502 \\
19 & 3 & 4 & 2 (N\_27) & 10 & 5\,521 & 639 & 2\,025 & 924 & 9\,109 \\ \addlinespace[2pt]
20 & 4 & 0 & 4 (N\_24) & 9 & 4\,633 & 1\,655 & 3\,410 & 924 & 10\,622 \\
21 & 4 & 1 & 0 (N\_65) & 8 & 5\,741 & 416 & 2\,237 & 924 & 9\,318 \\
22 & 4 & 2 & 1 (O\_51) & 9 & 5\,145 & 961 & 2\,536 & 924 & 9\,566 \\
23 & 4 & 3 & 2 (N\_27) & 12 & 5\,974 & 127 & 2\,537 & 924 & 9\,562 \\
24 & 4 & 4 & 3 (N\_81) & 10 & 4\,900 & 1\,260 & 5\,440 & 924 & 12\,524 \\
\bottomrule
\end{tabular}
\end{table}

\clearpage

\subsection{Statistical validity of the comparison of atomic models}
\label{app:cv_diagnostics}

The statistical validity of the model comparison based on a fixed holdout design was assessed through three diagnostics computed exclusively over atoms carrying the held-out cluster labels H\_10, C\_11, N\_13, and O\_10 — enforcing the condition that no metric is diluted by environments seen during training (see text and Tabs. \ref{tab:cv_diagnostics_cluster_general} and \ref{tab:cv_diagnostics_cluster_nonaug}). Because the 5 repeat means reflect only training-sample variability (not holdout-selection variability), Tukey's honest significant difference (HSD) error term is tighter than it would be under a random-holdout design. The resulting model rankings are statistically reliable for the defined OOD environments \{H\_10, C\_11, N\_13, O\_10\} and stable across training-set perturbations, as confirmed by the near-zero ICC. However, the confidence intervals do not capture uncertainty about the choice of holdout, so rankings should be interpreted as conditional on this specific evaluation set and not generalized to arbitrary OOD regimes.

\textbf{Fold independence (ICC).} The intraclass correlation ICC$(1,1)$ across the five inner folds within each repeat measures whether training-set composition variation causes correlated test performance within a repeat. All models returned near-zero or mildly negative ICC values when averaging over atoms and properties, indicating that folds contribute essentially independent information despite the partial overlap in training molecules. Negative values — observed in several strata of the SI table (Tab. \ref{tab:cv_diagnostics_cluster_full}) — arise naturally when training-set variability is dominated by which environments are \textit{excluded} rather than included, and do not indicate a violation. The effective sample size $n_\mathrm{eff} \in [16.5, 25.0]$ across models reflects this near-independence and confirms that the standard repeated-CV estimator remains close to unbiased. Sphericity corrections (Mauchly / Greenhouse–Geisser) are therefore not needed.

\textbf{Normality and homoscedasticity (SW and Levene).} Shapiro–Wilk normality tests on the five repeat-level means passed for 81.2–100\% when averaging over atoms and properties, only descending below 75\% when looking at per-atom type scores. Levene homoscedasticity tests passed for 93.8\% when scores were averaged over property and atom label, only descending to 75\% for certain atom types. Together, these diagnostics confirm that the RM-ANOVA and Tukey HSD assumptions are met in practice for CCC.

\begin{table}[h!]
  \caption{ANOVA validity diagnostics for averaged CCC metrics of rotationally augmented and equivariant models. All quantities computed exclusively over atoms carrying held-out cluster label. \textbf{ICC$(1,1)$}: intraclass correlation across the five inner folds within each repeat; ideal range $[-0.10,\,0.10]$ (near-independent folds, $n_{\text{eff}} \approx 25$); orange if $|\text{ICC}|>0.20$, red if $>0.40$. \textbf{$n_{\text{eff}}$}: effective sample size $= 25\,/\,(1+(k-1)\cdot\max(0,\text{ICC}))$; orange if ${<}80\%$ of nominal (${<}20$), red if ${<}60\%$ (${<}15$). \textbf{SW}: Shapiro--Wilk pass rate (\%) on five repeat-means; orange if ${<}80\%$, red if ${<}60\%$. \textbf{Lev.}: Levene homoscedasticity pass rate (\%) across models; same thresholds.}
  \label{tab:cv_diagnostics_cluster_general}
  \centering
  \begin{tabular}{lccccc}
    \toprule
    Diagnostic & \textbf{EGNN} & \textbf{SG-8-5} & \textbf{SG-8-12} & \textbf{SFC2} & \textbf{EGFC} \\
    \midrule
    ICC$(1,1)$ & 0.008\,$\pm$\,0.097 & 0.043\,$\pm$\,0.137 & -0.006\,$\pm$\,0.126 & 0.008\,$\pm$\,0.140 & -0.050\,$\pm$\,0.138 \\
    $n_\text{eff}$ & 22.3\,$\pm$\,3.7 & 20.2\,$\pm$\,4.7 & 22.0\,$\pm$\,4.4 & 21.5\,$\pm$\,4.5 & 22.7\,$\pm$\,4.0 \\
    SW & 100.0 & 87.5 & 81.2 & 100.0 & 100.0 \\
    Lev. & 93.8 & 93.8 & 93.8 & 93.8 & 93.8 \\
    \bottomrule
  \end{tabular}
\end{table}

\begin{table}[h!]
  \caption{ANOVA validity diagnostics for averaged CCC metrics of non-augmented models. All quantities computed exclusively over atoms carrying held-out cluster label. \textbf{ICC$(1,1)$}: intraclass correlation across the five inner folds within each repeat; ideal range $[-0.10,\,0.10]$ (near-independent folds, $n_{\text{eff}} \approx 25$); orange if $|\text{ICC}|>0.20$, red if $>0.40$. \textbf{$n_{\text{eff}}$}: effective sample size $= 25\,/\,(1+(k-1)\cdot\max(0,\text{ICC}))$; orange if ${<}80\%$ of nominal (${<}20$), red if ${<}60\%$ (${<}15$). \textbf{SW}: Shapiro--Wilk pass rate (\%) on five repeat-means; orange if ${<}80\%$, red if ${<}60\%$. \textbf{Lev.}: Levene homoscedasticity pass rate (\%) across models; same thresholds.}
  \label{tab:cv_diagnostics_cluster_nonaug}
  \centering
  \begin{tabular}{lcc}
    \toprule
    Diagnostic & \textbf{SGNN}  & \textbf{SGFC}  \\
    \midrule
    ICC$(1,1)$ &        -0.035\,$\pm$\,0.129             &  -0.116\,$\pm$\,0.064  \\
    $n_\text{eff}$ &    \cellcolor{orange!25}18.0\,$\pm$\,3.6 &  24.8\,$\pm$\,0.7  \\
    SW &                93.8 &  93.8  \\
    Lev. &              100.0 &  100.0  \\
    \bottomrule
  \end{tabular}
\end{table}

\newpage

\begin{sidewaystable}
  \caption{Full ANOVA validity diagnostics of the models in the main discussion per held-out label and property, CCC metrics. All quantities computed exclusively over atoms carrying held-out cluster label. \textbf{ICC$(1,1)$}: intraclass correlation across the five inner folds within each repeat; ideal range $[-0.10,\,0.10]$ (near-independent folds, $n_{\text{eff}} \approx 25$); orange if $|\text{ICC}|>0.20$, red if $>0.40$. \textbf{$n_{\text{eff}}$}: effective sample size $= 25\,/\,(1+(k-1)\cdot\max(0,\text{ICC}))$; orange if ${<}80\%$ of nominal (${<}20$), red if ${<}60\%$ (${<}15$). \textbf{SW}: Shapiro--Wilk pass rate (\%) on five repeat-means; orange if ${<}80\%$, red if ${<}60\%$. \textbf{Lev.}: Levene homoscedasticity pass rate (\%) across models; same thresholds.}
  \label{tab:cv_diagnostics_cluster_full}
  \centering
  \begin{tabular}{llcccccccccccccccc}
    \toprule
    \multicolumn{2}{l}{} & \multicolumn{4}{c}{H\_10} & \multicolumn{4}{c}{C\_11} & \multicolumn{4}{c}{N\_13} & \multicolumn{4}{c}{O\_10} \\
    \cmidrule(lr){3-6}\cmidrule(lr){7-10}\cmidrule(lr){11-14}\cmidrule(lr){15-18}
    Model & Property & ICC$(1,1)$ & $n_\text{eff}$ & SW & Lev. & ICC$(1,1)$ & $n_\text{eff}$ & SW & Lev. & ICC$(1,1)$ & $n_\text{eff}$ & SW & Lev. & ICC$(1,1)$ & $n_\text{eff}$ & SW & Lev. \\
    \midrule
    \multirow{4}{*}{\textbf{EGNN}} &$N$& -0.037 & 25.0 & 100.0 & 100.0 & -0.025 & 25.0 & 100.0 & 100.0 & -0.100 & 25.0 & 100.0 & 100.0 & 0.153 & \cellcolor{orange!25}15.5 & 100.0 & 100.0 \\
     &$\lambda$& 0.076 & \cellcolor{orange!25}19.2 & 100.0 & 100.0 & 0.020 & 23.1 & 100.0 & 100.0 & \cellcolor{orange!25}-0.205 & 25.0 & 100.0 & 100.0 & \cellcolor{orange!25}0.209 & \cellcolor{red!20}13.6 & 100.0 & \cellcolor{red!20}0.0 \\
     & $|\boldsymbol{\mu}|$ & -0.019 & 25.0 & 100.0 & 100.0 & -0.016 & 25.0 & 100.0 & 100.0 & -0.036 & 25.0 & 100.0 & 100.0 & 0.111 & \cellcolor{orange!25}17.3 & 100.0 & 100.0 \\
     & $|\mathbf{Q}|$ & 0.001 & 24.9 & 100.0 & 100.0 & 0.042 & 21.4 & 100.0 & 100.0 & -0.081 & 25.0 & 100.0 & 100.0 & 0.035 & 21.9 & 100.0 & 100.0 \\
    \addlinespace[4pt]
    \multirow{4}{*}{\textbf{SG-8-5}} &$N$& -0.071 & 25.0 & 100.0 & 100.0 & 0.070 & \cellcolor{orange!25}19.5 & 100.0 & 100.0 & 0.082 & \cellcolor{orange!25}18.8 & 100.0 & 100.0 & -0.017 & 25.0 & 100.0 & 100.0 \\
     &$\lambda$& -0.149 & 25.0 & 100.0 & 100.0 & 0.092 & \cellcolor{orange!25}18.3 & 100.0 & 100.0 & -0.132 & 25.0 & 100.0 & 100.0 & 0.129 & \cellcolor{orange!25}16.5 & \cellcolor{red!20}0.0 & \cellcolor{red!20}0.0 \\
     & $|\boldsymbol{\mu}|$ & -0.065 & 25.0 & 100.0 & 100.0 & 0.102 & \cellcolor{orange!25}17.7 & 100.0 & 100.0 & 0.019 & 23.2 & 100.0 & 100.0 & \cellcolor{orange!25}0.343 & \cellcolor{red!20}10.5 & 100.0 & 100.0 \\
     & $|\mathbf{Q}|$ & -0.177 & 25.0 & 100.0 & 100.0 & 0.061 & 20.1 & \cellcolor{red!20}0.0 & 100.0 & \cellcolor{orange!25}0.201 & \cellcolor{red!20}13.8 & 100.0 & 100.0 & 0.199 & \cellcolor{red!20}13.9 & 100.0 & 100.0 \\
    \addlinespace[4pt]
    \multirow{4}{*}{\textbf{SG-8-12}} &$N$& 0.049 & 20.9 & 100.0 & 100.0 & -0.011 & 25.0 & 100.0 & 100.0 & -0.143 & 25.0 & 100.0 & 100.0 & -0.165 & 25.0 & 100.0 & 100.0 \\
     &$\lambda$& \cellcolor{orange!25}0.213 & \cellcolor{red!20}13.5 & 100.0 & 100.0 & -0.100 & 25.0 & 100.0 & 100.0 & -0.066 & 25.0 & 100.0 & 100.0 & -0.196 & 25.0 & \cellcolor{red!20}0.0 & \cellcolor{red!20}0.0 \\
     & $|\boldsymbol{\mu}|$ & 0.146 & \cellcolor{orange!25}15.8 & 100.0 & 100.0 & 0.045 & 21.2 & 100.0 & 100.0 & 0.079 & \cellcolor{orange!25}19.0 & 100.0 & 100.0 & \cellcolor{orange!25}0.255 & \cellcolor{red!20}12.4 & 100.0 & 100.0 \\
     & $|\mathbf{Q}|$ & -0.073 & 25.0 & 100.0 & 100.0 & 0.002 & 24.9 & \cellcolor{red!20}0.0 & 100.0 & -0.055 & 25.0 & 100.0 & 100.0 & -0.068 & 25.0 & \cellcolor{red!20}0.0 & 100.0 \\
    \addlinespace[4pt]
    \multirow{4}{*}{\textbf{SFC2}} &$N$& 0.011 & 24.0 & 100.0 & 100.0 & -0.098 & 25.0 & 100.0 & 100.0 & 0.143 & \cellcolor{orange!25}15.9 & 100.0 & 100.0 & -0.042 & 25.0 & 100.0 & 100.0 \\
     &$\lambda$& 0.072 & \cellcolor{orange!25}19.4 & 100.0 & 100.0 & 0.001 & 24.9 & 100.0 & 100.0 & \cellcolor{orange!25}0.348 & \cellcolor{red!20}10.5 & 100.0 & 100.0 & 0.025 & 22.7 & 100.0 & \cellcolor{red!20}0.0 \\
     & $|\boldsymbol{\mu}|$ & 0.157 & \cellcolor{orange!25}15.3 & 100.0 & 100.0 & -0.160 & 25.0 & 100.0 & 100.0 & 0.069 & \cellcolor{orange!25}19.6 & 100.0 & 100.0 & 0.124 & \cellcolor{orange!25}16.7 & 100.0 & 100.0 \\
     & $|\mathbf{Q}|$ & -0.193 & 25.0 & 100.0 & 100.0 & -0.109 & 25.0 & 100.0 & 100.0 & -0.043 & 25.0 & 100.0 & 100.0 & -0.181 & 25.0 & 100.0 & 100.0 \\
    \addlinespace[4pt]
    \multirow{4}{*}{\textbf{EGFC}} &$N$& -0.104 & 25.0 & 100.0 & 100.0 & 0.007 & 24.3 & 100.0 & 100.0 & -0.127 & 25.0 & 100.0 & 100.0 & 0.050 & 20.8 & 100.0 & 100.0 \\
     &$\lambda$& -0.097 & 25.0 & 100.0 & 100.0 & 0.149 & \cellcolor{orange!25}15.7 & 100.0 & 100.0 & \cellcolor{orange!25}-0.208 & 25.0 & 100.0 & 100.0 & -0.090 & 25.0 & 100.0 & \cellcolor{red!20}0.0 \\
     & $|\boldsymbol{\mu}|$ & -0.178 & 25.0 & 100.0 & 100.0 & -0.178 & 25.0 & 100.0 & 100.0 & 0.093 & \cellcolor{orange!25}18.2 & 100.0 & 100.0 & 0.017 & 23.4 & 100.0 & 100.0 \\
     & $|\mathbf{Q}|$ & -0.195 & 25.0 & 100.0 & 100.0 & \cellcolor{orange!25}0.299 & \cellcolor{red!20}11.4 & 100.0 & 100.0 & -0.154 & 25.0 & 100.0 & 100.0 & -0.086 & 25.0 & 100.0 & 100.0 \\
    \bottomrule
  \end{tabular}
\end{sidewaystable}

\clearpage

\subsection{Analysis of environment leakage due to model cutoff}
\label{SI:cutoff_leakage}

The out-of-distribution holdout — the environment clusters \{H\_10, C\_11, N\_13, O\_10\} — was defined by clustering SOAP descriptors computed at a cutoff of 4.5 Bohr. The atomic models, however, build their message-passing graph at a larger cutoff (5.25 Bohr for EGNN, 8 Bohr for QT-Net), and it is the cutoff that sets the graph topology, i.e. which atoms a central atom perceives. A held-out environment that is genuinely distinct from training at 4.5 Bohr may cease to be distinct once the neighborhood is widened, because the larger sphere pulls in context that makes it resemble a training environment. If that happens, the held-out set is less out-of-distribution at the model's radius than its 4.5-Bohr construction implies — it leaks. This appendix quantifies that effect with a single, clustering-free, calibrated measure of novelty, evaluated as a function of the cutoff $r_c$.

For each cutoff $r_c$ and each element we recompute every atom's SOAP descriptor and project it onto the principal components of the full dataset, keeping enough components to retain the same per-element variance the original 4.5-Bohr projection retained (O 99.04\%, H 99.11\%, C 99.01\%, N 99.18\%); at 4.5 Bohr we reuse the exact per-element component counts of the original clustering (H 20, C 35, N 30, O 20). Holding the information budget fixed in this way ensures the only quantity varying along the sweep is the radius, not how much of the descriptor is retained. 

We then rescale each principal component by the standard deviation of the \textit{training} atoms along it, so that every direction contributes to distance in proportion to how much the training set actually varies along it (equivalently, Euclidean distance in the rescaled space is approximately the Mahalanobis distance with respect to the training covariance). For each held-out (query) atom we compute the distance $d_{15}$ to its 15th nearest training neighbor. Since $d_{15}$ is in arbitrary units that differ by element and radius, we split the training atoms into two disjoint groups: a reference set $R$ ($\leq$ 150\,000 atoms) and a calibration set C ($\leq$100\,000 atoms). The distances from the calibration atoms to
$R$ give a reference distribution for atoms we know are training-typical, against which each held-out atom's distance to the same $R$ is ranked. A held-out atom's novelty percentile is then the percentage of calibration atoms with smaller $d_{15}$: a value of 50 means the atom is as isolated as the median training atom (it sits inside training's typical density and is effectively seen), whereas a value near 100 means it lies in a tail that training barely covers (genuinely OOD). 

\begin{figure}[h!]
    \centering
    \includegraphics[width=0.7\linewidth]{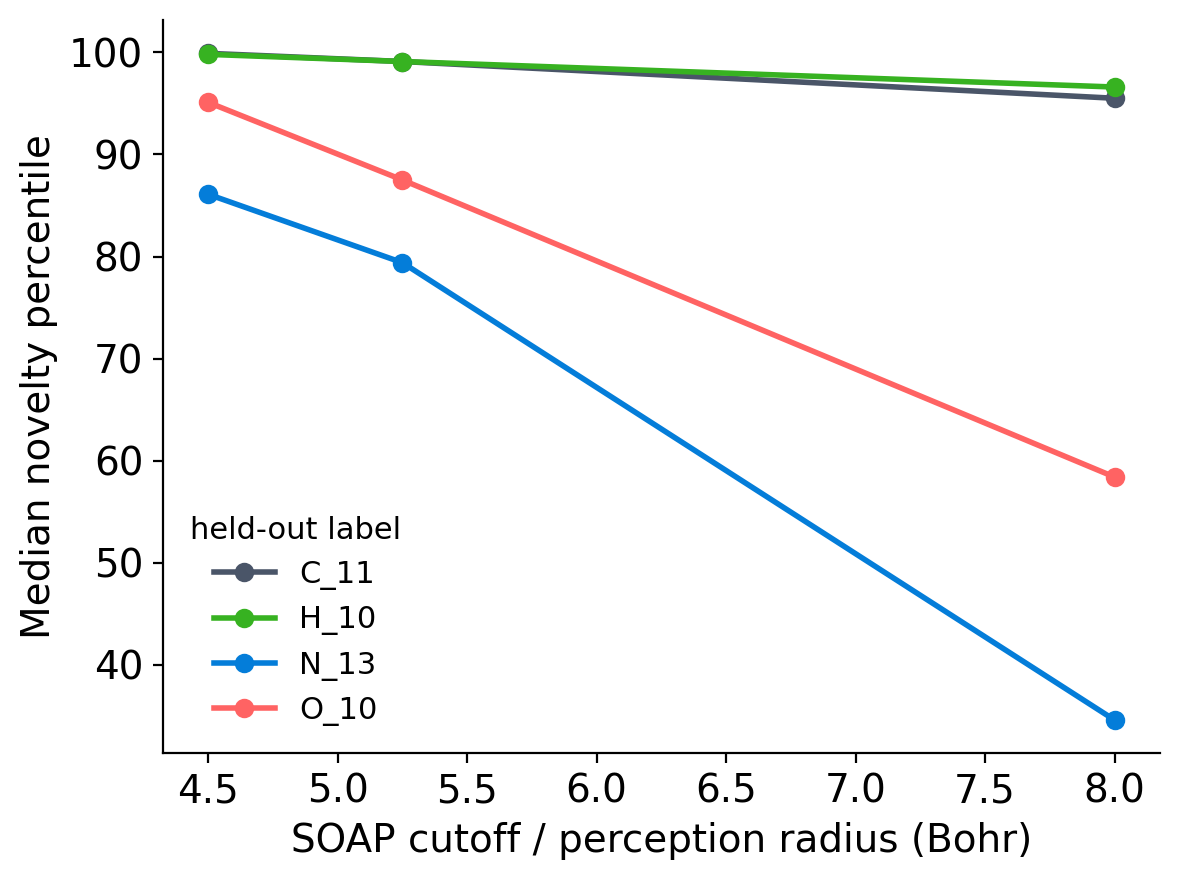}
    \caption{Median novelty percentile of the held-out environments at 4.5, 5.25 and 8 Bohr.}
    \label{fig:leakage_vs_radius}
\end{figure}

In Figure \ref{fig:leakage_vs_radius} we show the median novelty percentile across the three radii. The carbon and hydrogen holdouts are essentially unaffected: their median novelty stays above 95 at every radius, they remain genuinely out-of-distribution at the models' perception radius. The nitrogen and oxygen holdouts behave very differently. As the cutoff widens from 4.5 to 8 Bohr, the median novelty of N\_13 falls from 86 to 35 and that of O\_10 from 95 to 58. To test whether this leakage is inflating the scores, we re-stratify the held-out N\_13 and O\_10 atoms at the 8-Bohr cutoff into a leaked subset (novelty percentile $< 50$) and a novel subset, and compute the mean CCC (see Tab. \ref{tab:novel_vs_leaked}). Crucially, the leaked atoms are predicted less accurately than those that remain novel. The reported OOD metrics are therefore not inflated by the cutoff mismatch, and the model-comparison conclusions — already framed as conditional on the chosen holdout — are unaffected.

\begin{table}
\caption{Mean CCC of each model for each property of the leaked and novel atoms.}
  \label{tab:novel_vs_leaked}
  \centering
  \begin{tabular}{@{}llcccc@{}}
\toprule
\multicolumn{2}{l}{}                                     & \multicolumn{2}{c}{N\_13} & \multicolumn{2}{c}{O\_10} \\ \cmidrule(l){3-6} 
Model                             & Property             & Leaked       & Novel      & Leaked       & Novel      \\ \midrule
\multirow{4}{*}{\textbf{EGNN}}    & $N$                  & 0.590        & 0.806      & 0.852        & 0.906      \\
                                  & $\lambda$            & 0.595        & 0.810      & 0.830        & 0.878      \\
                                  & $|\boldsymbol{\mu}|$ & 0.526        & 0.569      & 0.921        & 0.927      \\
                                  & $|\mathbf{Q}|$       & 0.380        & 0.603      & 0.856        & 0.854      \\
\multirow{4}{*}{\textbf{SG-8-5}}  & $N$                  & 0.839        & 0.919      & 0.941        & 0.958      \\
                                  & $\lambda$            & 0.832        & 0.904      & 0.949        & 0.954      \\
                                  & $|\boldsymbol{\mu}|$ & 0.683        & 0.776      & 0.928        & 0.946      \\
                                  & $|\mathbf{Q}|$       & 0.654        & 0.852      & 0.869        & 0.893      \\
\multirow{4}{*}{\textbf{SG-8-12}} & $N$                  & 0.884        & 0.961      & 0.949        & 0.967      \\
                                  & $\lambda$            & 0.897        & 0.955      & 0.961        & 0.971      \\
                                  & $|\boldsymbol{\mu}|$ & 0.821        & 0.888      & 0.954        & 0.971      \\
                                  & $|\mathbf{Q}|$       & 0.675        & 0.889      & 0.902        & 0.924      \\
\multirow{4}{*}{\textbf{SFC2}}    & $N$                  & 0.852        & 0.955      & 0.958        & 0.974      \\
                                  & $\lambda$            & 0.879        & 0.946      & 0.964        & 0.975      \\
                                  & $|\boldsymbol{\mu}|$ & 0.818        & 0.898      & 0.951        & 0.968      \\
                                  & $|\mathbf{Q}|$       & 0.767        & 0.921      & 0.889        & 0.907      \\
\multirow{4}{*}{\textbf{EGFC}}    & $N$                  & 0.654        & 0.853      & 0.917        & 0.941      \\
                                  & $\lambda$            & 0.715        & 0.859      & 0.925        & 0.938      \\
                                  & $|\boldsymbol{\mu}|$ & 0.634        & 0.724      & 0.846        & 0.880      \\
& $|\mathbf{Q}|$                    & 0.457                & 0.650        & 0.683      & 0.727 \\
\bottomrule
\end{tabular}
\end{table}

\clearpage

\subsection{Additional plots and metrics for the discussion on predicting QTA properties}\label{SI:test}

\begin{figure}[h]
    \centering
    \includegraphics[width=1.0\linewidth]{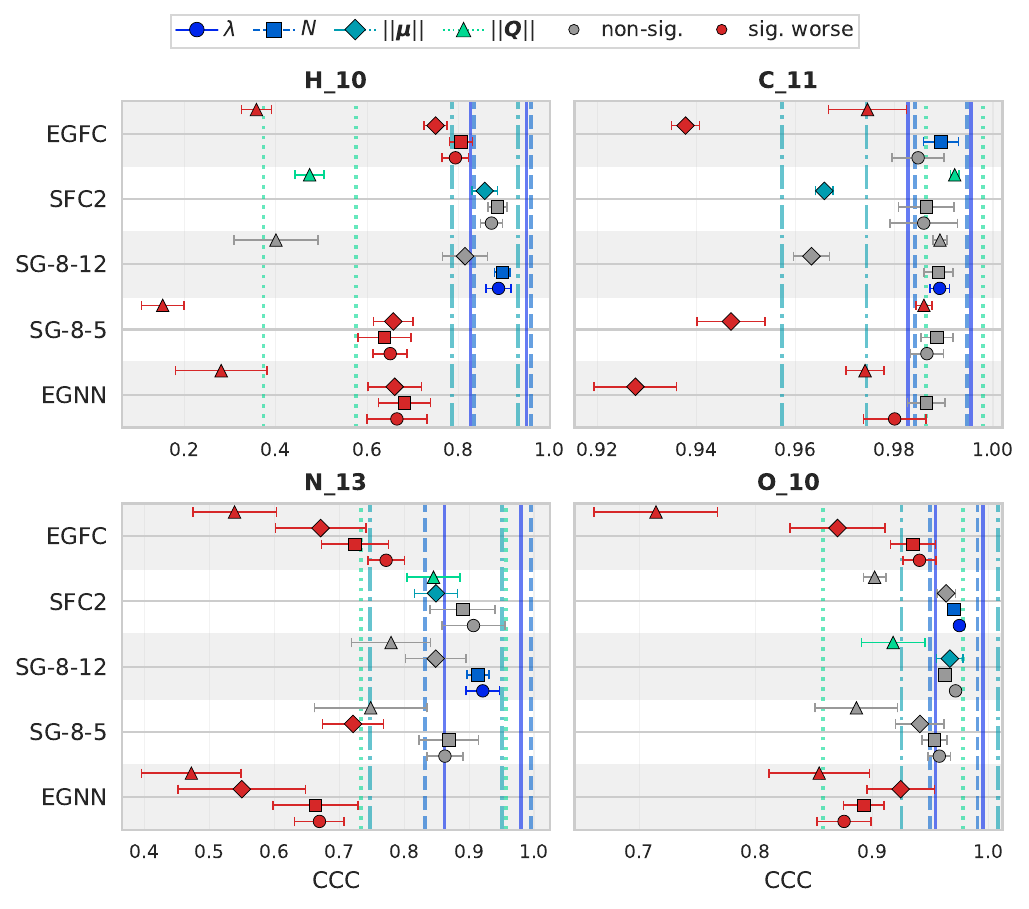}
    \caption{Tukey forest plots for CCC scores of each model, for each atom and property in the holdout set. The best performing model for each (atom, property) shows a colored symbol surrounded by the minimum significant difference.}
    \label{fig:Tukey_clusters_CCC}
\end{figure}

\begin{figure}[h]
    \centering
    \includegraphics[width=1.0\linewidth]{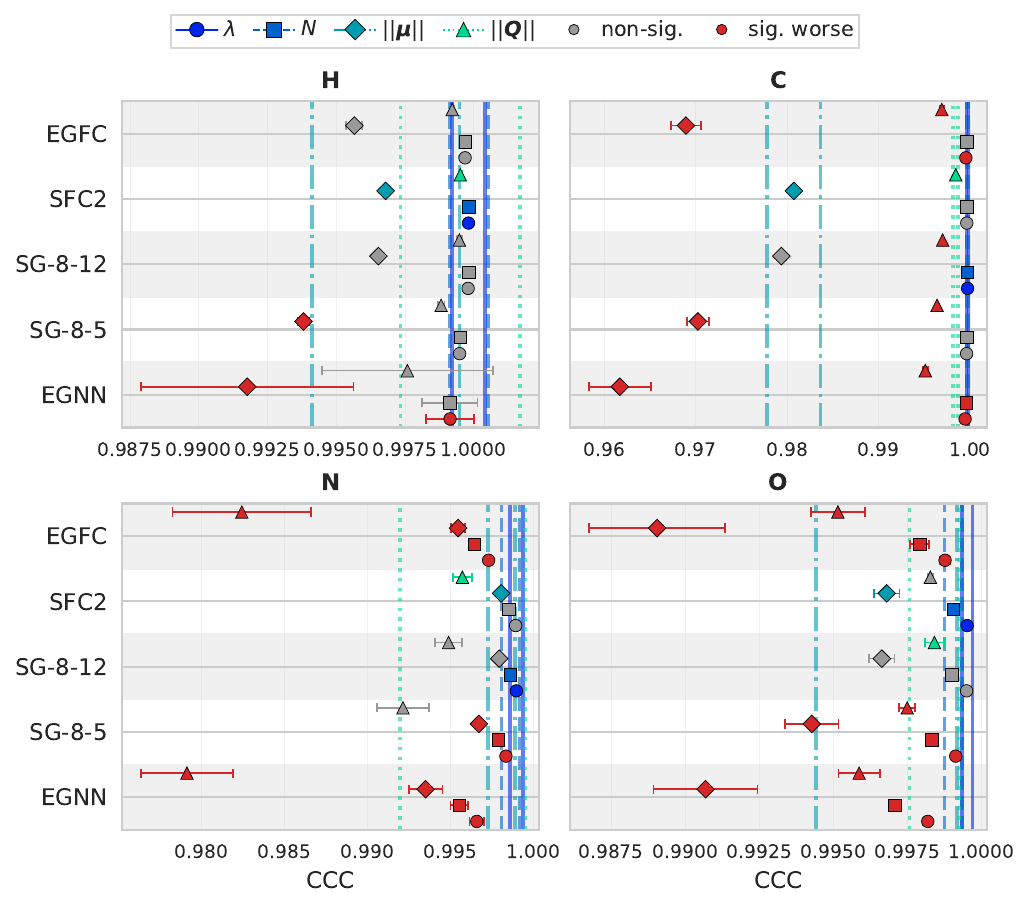}
    \caption{Tukey forest plots for CCC scores of each model, for each atom and property for all of the test set. The best performing model for each (atom, property) shows a colored symbol surrounded by the minimum significant difference.}
    \label{fig:Tukey_test_CCC}
\end{figure}

We include here numerical values of CCC (Tab. \ref{tab:metrics}) and Tukey forest plots for the equivariant and rotationally augmented models (Fig.~\ref{fig:Tukey_clusters_CCC}). Per-element and per-property differences must be interpreted keeping in mind that the statistical validity gets compromised at this stratum (see Tab. \ref{tab:cv_diagnostics_cluster_full}). It is worth noticing that, when accounting for all atoms in the test set, all models achieve CCC > 0.9 in all properties and atoms (Fig. \ref{fig:Tukey_test_CCC}).

\newpage
\begin{sidewaystable}
  \caption{Model comparison on CCC (mean\,$\pm$\,95\,\%\,CI over $n=5$ CV repeats). Cell color: \colorbox{gold!50}{1st}\,\colorbox{gray!25}{2nd}\,\colorbox{orange!30}{3rd} per row.}
  \label{tab:metrics}
  \centering
  \begin{tabular}{llccccc}
    \toprule
    Cluster & Property & \textbf{EGNN} & \textbf{SG-8-5} & \textbf{SG-8-12} & \textbf{SFC2} & \textbf{EGFC} \\
    \midrule
    \multirow{4}{*}{H\_10} &$N$& 0.682\,$\pm$\,0.057 & 0.638\,$\pm$\,0.058 & \cellcolor{gold!50}0.897\,$\pm$\,0.017 & \cellcolor{gray!25}0.886\,$\pm$\,0.021 & 0.806\,$\pm$\,0.025 \\
     &$\lambda$& 0.665\,$\pm$\,0.066 & 0.651\,$\pm$\,0.037 & \cellcolor{gold!50}0.888\,$\pm$\,0.028 & \cellcolor{gray!25}0.873\,$\pm$\,0.024 & 0.794\,$\pm$\,0.029 \\
     & $|\boldsymbol{\mu}|$ & 0.661\,$\pm$\,0.059 & 0.658\,$\pm$\,0.044 & \cellcolor{gray!25}0.815\,$\pm$\,0.049 & \cellcolor{gold!50}0.858\,$\pm$\,0.028 & \cellcolor{orange!30}0.750\,$\pm$\,0.025 \\
     & $|\mathbf{Q}|$ & 0.280\,$\pm$\,0.100 & 0.152\,$\pm$\,0.046 & \cellcolor{gray!25}0.400\,$\pm$\,0.093 & \cellcolor{gold!50}0.474\,$\pm$\,0.032 & \cellcolor{orange!30}0.357\,$\pm$\,0.033 \\
    \addlinespace[2pt]
    \multirow{4}{*}{C\_11} &$N$& 0.987\,$\pm$\,0.004 & \cellcolor{orange!30}0.989\,$\pm$\,0.003 & \cellcolor{gray!25}0.989\,$\pm$\,0.003 & 0.986\,$\pm$\,0.006 & \cellcolor{gold!50}0.989\,$\pm$\,0.004 \\
     &$\lambda$& 0.980\,$\pm$\,0.006 & \cellcolor{gray!25}0.987\,$\pm$\,0.003 & \cellcolor{gold!50}0.989\,$\pm$\,0.002 & \cellcolor{orange!30}0.986\,$\pm$\,0.007 & 0.985\,$\pm$\,0.005 \\
     & $|\boldsymbol{\mu}|$ & 0.928\,$\pm$\,0.008 & 0.947\,$\pm$\,0.007 & \cellcolor{gray!25}0.963\,$\pm$\,0.004 & \cellcolor{gold!50}0.966\,$\pm$\,0.002 & 0.938\,$\pm$\,0.003 \\
     & $|\mathbf{Q}|$ & 0.974\,$\pm$\,0.004 & \cellcolor{orange!30}0.986\,$\pm$\,0.002 & \cellcolor{gray!25}0.989\,$\pm$\,0.001 & \cellcolor{gold!50}0.992\,$\pm$\,0.001 & 0.975\,$\pm$\,0.008 \\
    \addlinespace[2pt]
    \multirow{4}{*}{N\_13} &$N$& 0.664\,$\pm$\,0.066 & \cellcolor{orange!30}0.869\,$\pm$\,0.046 & \cellcolor{gold!50}0.914\,$\pm$\,0.017 & \cellcolor{gray!25}0.890\,$\pm$\,0.050 & 0.724\,$\pm$\,0.051 \\
     &$\lambda$& 0.670\,$\pm$\,0.038 & \cellcolor{orange!30}0.863\,$\pm$\,0.028 & \cellcolor{gold!50}0.921\,$\pm$\,0.026 & \cellcolor{gray!25}0.907\,$\pm$\,0.048 & 0.772\,$\pm$\,0.028 \\
     & $|\boldsymbol{\mu}|$ & 0.550\,$\pm$\,0.098 & 0.721\,$\pm$\,0.047 & \cellcolor{gray!25}0.849\,$\pm$\,0.046 & \cellcolor{gold!50}0.849\,$\pm$\,0.033 & 0.671\,$\pm$\,0.069 \\
     & $|\mathbf{Q}|$ & 0.473\,$\pm$\,0.077 & \cellcolor{orange!30}0.749\,$\pm$\,0.086 & \cellcolor{gray!25}0.780\,$\pm$\,0.061 & \cellcolor{gold!50}0.845\,$\pm$\,0.040 & 0.539\,$\pm$\,0.064 \\
    \addlinespace[2pt]
    \multirow{4}{*}{O\_10} &$N$& 0.893\,$\pm$\,0.017 & \cellcolor{orange!30}0.954\,$\pm$\,0.011 & \cellcolor{gray!25}0.963\,$\pm$\,0.003 & \cellcolor{gold!50}0.970\,$\pm$\,0.005 & 0.935\,$\pm$\,0.019 \\
     &$\lambda$& 0.876\,$\pm$\,0.023 & \cellcolor{orange!30}0.958\,$\pm$\,0.010 & \cellcolor{gray!25}0.972\,$\pm$\,0.002 & \cellcolor{gold!50}0.975\,$\pm$\,0.004 & 0.941\,$\pm$\,0.014 \\
     & $|\boldsymbol{\mu}|$ & 0.925\,$\pm$\,0.029 & \cellcolor{orange!30}0.941\,$\pm$\,0.021 & \cellcolor{gold!50}0.967\,$\pm$\,0.011 & \cellcolor{gray!25}0.964\,$\pm$\,0.008 & 0.870\,$\pm$\,0.041 \\
     & $|\mathbf{Q}|$ & 0.855\,$\pm$\,0.043 & \cellcolor{orange!30}0.887\,$\pm$\,0.035 & \cellcolor{gold!50}0.918\,$\pm$\,0.027 & \cellcolor{gray!25}0.902\,$\pm$\,0.010 & 0.715\,$\pm$\,0.053 \\
    \bottomrule
  \end{tabular}
\end{sidewaystable}

\clearpage

\subsection{Additional discussion on the effect of layer depth}\label{SI:augment}

\begin{figure}[h!]
    \centering
    \includegraphics[width=\linewidth]{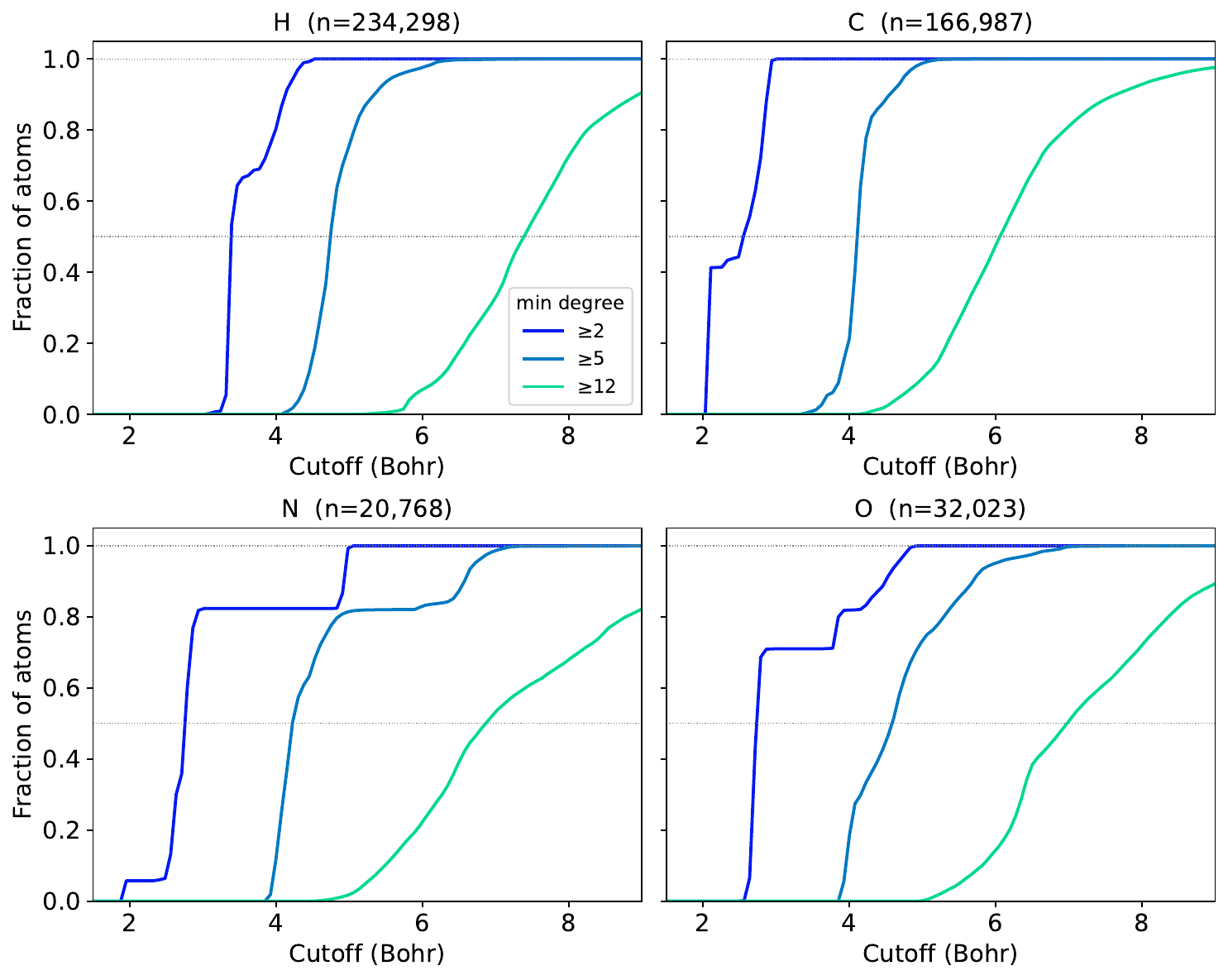}
    \caption{Distributions of fractions of atoms in training by minimum number of neighbors at each cutoff.}
    \label{fig:rdfs_training}
\end{figure}

Here, we provide arguments in favor that the greater depth of the scalar models is not being invested in passing chemical information, but rather in polishing the purely geometrical information that equivariant models understand by design. To confirm this, we trained two non-augmented, scalar GNNs matching the connectivity of EGNN and EGFC. We refer to these two non-augmented models as SGNN and SGFC. In Figure \ref{fig:rdfs_training}, we include radial distributions for reference.

The results in Tab. \ref{tab:p-value_non-aug} show that the difference between EGNN and SGNN is not significant. Therefore, at equal connectivity and without data augmentation, the greater depth is being invested in geometric learning. The significant performance difference between SGFC and EGFC suggests that the geometric understanding of non-equivariant models can highly benefit from a denser connectivity.

\begin{table}[h]
\caption{Tukey p-value matrix of CCC scores averaged over atom types and properties, comparing non-data augmented, non-equivariant models with equivariant models. A value below 0.05 indicates a statistically significant difference in performance. The best performing models are determined based on CCC mean value and statistical significance of the difference: \colorbox{gold!50}{1st}\,\colorbox{gray!25}{2nd}\,\colorbox{orange!30}{3rd}.}
\label{tab:p-value_non-aug}
\centering
\begin{tabular}{@{}lccccccc@{}}
\toprule
                           & \multicolumn{4}{c}{Tukey p-values}                    &                                    \\ \cmidrule(lr){2-5}
\multicolumn{1}{c}{Models} & EGFC   & SGFC    &EGNN   & SGNN   & CCC ± 95\% CI             \\ \midrule
EGFC                       & 1.0000 & 0.0010  &0.0010 & 0.0010 & \colorbox{gray!25}{0.798$\pm$0.006} \\
SGFC                       & 0.0010 & 1.0000  &0.0010 & 0.0010 & \colorbox{gold!50}{0.844$\pm$0.011} \\
EGNN                       & 0.0010 & 0.0010  &1.0000 & \textbf{0.6128} & \colorbox{orange!30}{0.754$\pm$0.028} \\
SGNN                       & 0.0010 & 0.0010  &\textbf{0.6128} & 1.0000 & \colorbox{orange!30}{0.744$\pm$0.018} \\ \bottomrule
\end{tabular}
\end{table}

\clearpage

\subsection{Additional plots and metrics for the discussion on molecular predictions}\label{SI:descriptors}\label{SI:dipole_errors}

This section complements Sec.~\ref{sec:inference} with the sources of error of the reconstructed dipoles, the comparison of informed and blind models, the method behind the same-molecule comparison of Fig.~\ref{fig:same_molecule}, and the error of the molecular models against the distance to their training sets.

\begin{figure}[h!]
    \centering
    \includegraphics[width=\linewidth]{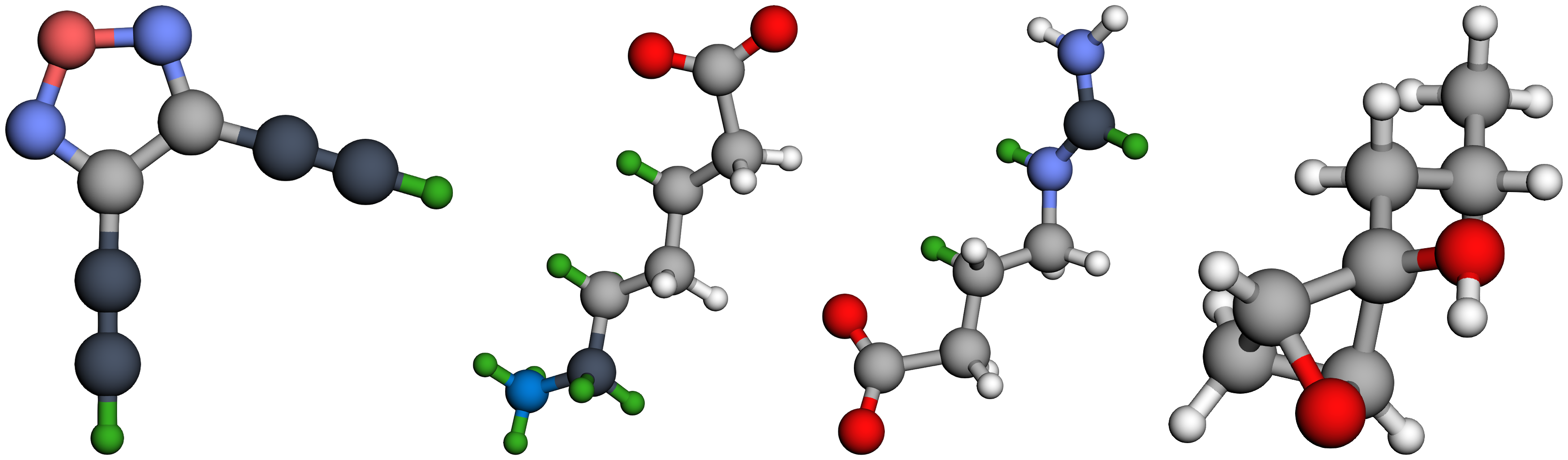}
    \caption{The molecule with the largest error of the reconstructed dipole in each group of Table~\ref{tab:dipole_groups}. From left to right: an alkyne (\texttt{C\#CC1=NON=C1C\#C}; QM9 dipole 3.8\,D vs reconstructed 0.6\,D, whereas most alkynes off by more than 1\,D are overestimated), a protonated zwitterion (\texttt{[NH3+]CCCCCC([O-])=O}; 29.6 vs 15.1\,D), another zwitterion (\texttt{NC=[NH+]CCCC([O-])=O}; 21.9 vs 14.5\,D) and an alkyl-oxabicyclopentanol (\texttt{CCCC1(O)C2CC1O2}; 1.7 vs 9.3\,D). Atoms within AIMEl's 99th percentile of the distance of Appendix~\ref{SI:QM9_clustering} are drawn in the colors of Fig.~\ref{fig:pipeline} for atoms in seen environments (\colorbox{Hcolor}{H},
\colorbox{Ccolor}{C},
\colorbox{Ncolor}{\textcolor{white}{N}},
\colorbox{Ocolor}{O}), atoms beyond it in its colors for held-out environments (\colorbox{H10color}{\textcolor{white}{H}},
\colorbox{C11color}{\textcolor{white}{C}},
\colorbox{N13color}{\textcolor{white}{N}},
\colorbox{O10color}{O}).}
    \label{fig:dipole_molecules}
\end{figure}

\textbf{Reconstructed dipoles.} Of the remainder's molecules, 15\% have three or more atoms far from AIMEl. They carry 73\% of the absolute error of the reconstructed dipoles (Fig.~\ref{fig:dipole_from_inferred}), with 99.3\% of the molecules that are off by more than 1\,D containing at least one rare atom. Two chemical groups explain most of this error (Table~\ref{tab:dipole_groups}, Fig.~\ref{fig:dipole_molecules}). Most of the contribution comes from alkynes: AIMEl contains only 14 molecules with a C$\equiv$C bond, whereas the remainder contains 17\,468. These carry 77\% of the absolute and squared error and represent 95\% of the molecules whose predictions are off by more than 1\,D. Zwitterions also contribute significantly, with 83\% of them containing a rare atom. They represent 0.54\% of the remainder but carry 86 of the 100 largest errors and 20\% of the squared error. The only large errors without rare atoms are four 4-alkyl-2-oxabicyclo[1.1.1]pentan-4-ols, off by 6.0--7.5\,D: AIMEl contains their strained cage in 527 molecules, but their substitution pattern just once. QT-Net's receptive field spans a whole QM9 molecule, so what is new to it is the molecule rather than a 6-Bohr environment. The groups also differ in the sign of their errors: among the molecules off by more than 1\,D, QT-Net underestimates the dipole of every zwitterion, and overestimates that of 77\% of the alkynes and of all the oxabicyclopentanols.

\begin{table}[h!]
\caption{Chemical groups behind the largest errors of the reconstructed dipoles on the remainder. The groups do not overlap: zwitterions with a C$\equiv$C bond count as zwitterions. Share: of the total absolute and squared error. Off: molecules with an absolute error above 1\,D. $R^2$ without: of the remainder without the group (0.931 with all molecules).}
\label{tab:dipole_groups}
\centering\small
\begin{tabular}{@{}lrrrrrrr@{}}
\toprule
 & \multicolumn{2}{c}{Molecules} & & \multicolumn{2}{c}{Share (\%)} & & \\ \cmidrule(lr){2-3}\cmidrule(lr){5-6}
Group & AIMEl & Remainder & MAE (D) & Abs. & Sq. & Off & $R^2$ without \\ \midrule
Alkynes & 14 & 17\,468 & 0.674 & 77.1 & 76.9 & 4\,888 & 0.981 \\
Protonated zwitterions & 13 & 195 & 2.468 & 3.2 & 16.1 & 125 & 0.934 \\
Other zwitterions & 37 & 332 & 0.904 & 2.0 & 4.0 & 103 & 0.929 \\
Alkyl-oxabicyclopentanols & 1 & 10 & 3.053 & 0.2 & 1.1 & 5 & 0.931 \\ \bottomrule
\end{tabular}
\end{table}

\textbf{Same molecules and distance to the training set.} The panels of a fold hold different molecules, and the molecules that carry withheld environments are harder to predict for any model. In consequence, comparing panels mixes the cost of a holdout with the difficulty of the molecules it selects. Figure~\ref{fig:same_molecule} instead compares each molecule with itself. For two panels X and Y that differ only in whether an atomic environment or a chemotype was seen in training (Appendix~\ref{SI:QM9_clustering}), the error increase in fold $k$ is
\begin{equation*}
\Delta_k^{X|Y} = \frac{100}{4}\sum_{p}\left(\frac{\sum_{i\in X_k} e^{p}_{ik}}{\sum_{i\in X_k} \bar e^{\,p}_{i,Y}} - 1\right),
\qquad e^{p}_{ik} = \frac{|\hat y^{p}_{ik} - y^{p}_{i}|}{\sigma_p},
\end{equation*}
where $\hat y^{p}_{ik}$ is the prediction of the model of fold $k$ for target $p$ of molecule $i$, $\sigma_p$ is the standard deviation of target $p$ over AIMEl, so that the four targets share one scale, $X_k$ holds the molecules of panel X in fold $k$ that sit in panel Y in other folds, and $\bar e^{\,p}_{i,Y}$ is the mean error of molecule $i$ over those folds. The ratio compares the total error of these molecules with the total error that the same molecules receive from the models that saw the environment or the chemotype, so the difficulty of each molecule cancels. Each factor has two such contrasts: unseen environments are B$|$A (with the chemotype withheld) and C$|$D (with the chemotype seen). An unseen chemotype is A$|$D (environments seen) and B$|$C (environments withheld). All four are positive at every training fraction on both test sets, and for each target separately on the AIMEl test sets. Figure~\ref{fig:same_molecule} shows the mean of the two contrasts of each factor over the 25 folds. On the QM9 remainder, every molecule is predicted by the 25 models of a fraction, which play the part of the folds, and its panel depends on each model's training set. The molecules of the two chemotypes absent from AIMEl are left out, since their chemotype is never seen.

\begin{figure}[h!]
    \centering
    \includegraphics[width=\linewidth]{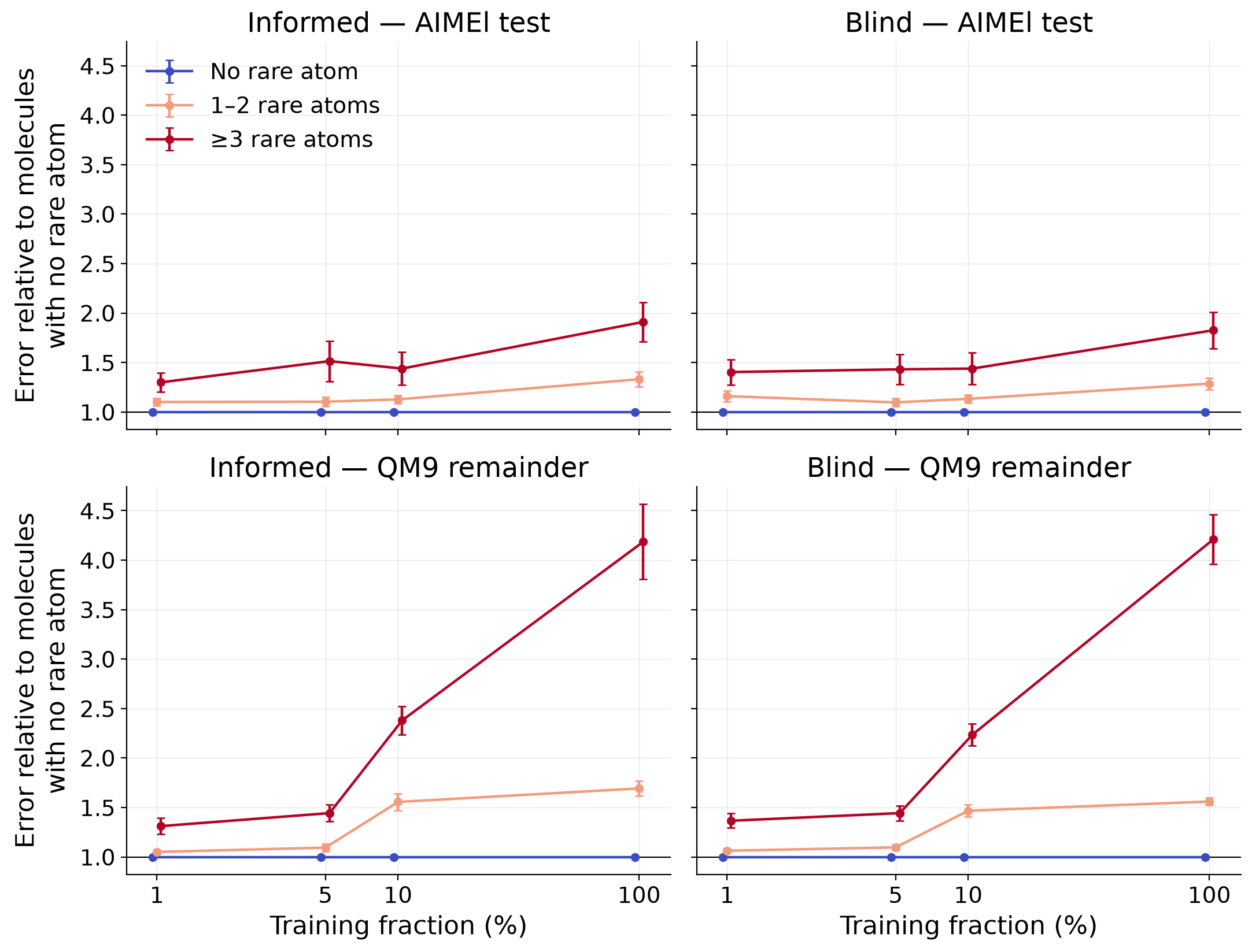}
    \caption{Error of test molecules with 1--2 (salmon) and $\geq$3 (red) atoms rare with respect to the model's training set, relative to the molecules with none (blue, at 1) for the same model, against the training fraction. Rows: AIMEl test sets (top) and 20\,000 remainder molecules (bottom); columns: informed (left) and blind (right) models. Fold means with 95\% $t$-intervals.}
    \label{fig:error_vs_novelty}
\end{figure}

The distance of Appendix~\ref{SI:QM9_clustering} grades how far each atom lies from a model's own training set. In every model configuration, the error increases with this distance. Molecules with $\geq$3 rare atoms have a larger error than molecules with none (Fig.~\ref{fig:error_vs_novelty}). The penalty grows with the training set, as in Fig.~\ref{fig:same_molecule}: on the AIMEl test sets, from 1.3--1.4 times the error of molecules without rare atoms with 160 training molecules to 1.8--1.9 times with the full training set, and up to 4.2 times on the remainder. The more a model learns from the environments it has seen, the more the environments it has not seen cost.

\clearpage

\subsection{Equivariant models}\label{SI:equivariant}

In the following, we use $L$ to denote the label of the irreducible representation (irrep) of $SO(3)$ with dimension $2L+1$. We represent each atom by a node with scalar features $h_i$, vector features $\mathbf{x}_i$ and traceless, symmetric features $\mathbf{T}_i$. For edge features, we follow the same convention, but with double subindices (e.g., $h_{ij}$). We denote the channel-wise (Hadamard) product with $\odot$ and the outer product with $\otimes$. We refer to the whole array of features by $f \equiv \left(h, \mathbf{x}, \mathbf{T}\right)$, and to the $L>0$ (non-scalar) features by $\mathbf{f} \equiv \left(\mathbf{x}, \mathbf{T}\right)$. We refer to the concatenation of invariants by $\|f\|\equiv \left(h,\left\|\mathbf{x}\right\|,\left\|\mathbf{T}\right\|\right)$. Throughout the models, we apply normalization separately to each irreducible representation: $\mathrm{LayerNorm}$ for scalar features and the equivariant $\mathrm{E3Norm}$ of \citet{batatia2023macehigherorderequivariant} for vector and tensor features. We denote normalized features uniformly by $\tilde{f} = \mathrm{E3Norm}(f)$, with the understanding that $\mathrm{E3Norm}$ reduces to $\mathrm{LayerNorm}$ for the scalar case. For non-scalar features,
\begin{equation*}
    \tilde{\mathbf{f}}_{i,c} = w_c\ \cdot\frac{\mathbf{f}_{i,c}}{\sqrt{\langle\|\mathbf{f}_{i,c}\|^2\rangle_c+\varepsilon}}, 
\end{equation*}
where $i$ is the cell (atom) index, $c$ indexes the parallel feature channels within a given irrep, $w_c$ is a learnable per-channel scaling weight, and $\varepsilon$ is a small constant for numerical stability. Because the normalization acts only on the magnitude of $\mathbf{f}_{i,c}$ and not on its components individually, the orientation of each vector or tensor feature is preserved — which is what allows $\mathrm{E3Norm}$ to be used inside an equivariant architecture without breaking the equivariance.

In all models, node features are initialized with a message passing layer as follows:
\begin{equation}
    h_i^{(0)} = \mathrm{LayerNorm}\left(\sum_j m^s_{ij} \odot \Gamma_0^s\left(r_{ij}\right)\right)
\end{equation}
\begin{equation}
    {\mathbf{x}_i}^{(0)} = \mathrm{E3Norm}\left(\sum_j m^v_{ij} \odot \Gamma_0^v\left(r_{ij}\right)\otimes\hat{\mathbf{r}}_{ij}\right)
\end{equation}
\begin{equation}
    {\mathbf{T}_i}^{(0)} = \mathrm{E3Norm}\left(\sum_j m^t_{ij} \odot \Gamma_0^t\left(r_{ij}\right)\otimes\hat{\mathbf{G}}_{ij}\right)
\end{equation}

where the messages $m_{ij}$ are generated by a multi-layer perceptron (MLP) that takes as input the embeddings of the atomic species, $m_{ij}=\mathrm{MLP}\left(\mathrm{EMBED}(Z_i),\mathrm{EMBED}(Z_j)\right)$, and $\Gamma^s$, $\Gamma^v$, $\Gamma^t$ are geometric filters that map a projection of the inter-atomic displacement onto a radial basis functions, and output through channel-wise sigmoid gates. 
Finally, $\hat{\mathbf{G}}_{ij}\propto\hat{\mathbf{r}}_{ij}\otimes\hat{\mathbf{r}}_{ij}$ is the unit-norm outer product of the relative position vector, expressed in the traceless symmetric representation as \begin{equation*}
    \hat{\mathbf{G}}(\hat{\mathbf{r}}) = \sqrt{3/2}\,[\,r_xr_y,\ r_xr_z,\ r_yr_z,\ (r_x^2-r_y^2)/2,\ r_z^2-1/3\,]\equiv[G_{xy},G_{xz},G_{yz},G_{an},G_{zz}].
\end{equation*}
The unit norm follows from the Frobenius norm in Eq. \ref{eq:frob_norm}.

Edge features are initialized from these node features as follows
\begin{equation}
    h_{ij}^{(0)} = \mathrm{LayerNorm}\left(\Psi^s\left(\mathrm{LayerNorm}\left(\sum_{k\in\{i,j\}}\psi^s\left(\|f_k^{(0)}\|\right)\right)\right)\odot \Gamma_1^s\left(r_{ij}\right)\right)
\end{equation}
\begin{equation}
    \mathbf{x}_{ij}^{(0)} = \mathrm{E3Norm}\left(\left(\sum_{k\in\{i,j\}}\psi^v\left(\|f_k^{(0)}\|\right)\odot\left(W^v\mathbf{x_k^{(0)}}\right)\right)\odot \Gamma_1^v\left(r_{ij}\right)\right)
\end{equation}
\begin{equation}
    \mathbf{T}_{ij}^{(0)} = \mathrm{E3Norm}\left(\left(\sum_{k\in\{i,j\}}\psi^t\left(\|f_k^{(0)}\|\right)\odot\left(W^t\mathbf{T_k^{(0)}}\right)\right)\odot \Gamma_1^t\left(r_{ij}\right)\right)
\end{equation}
where the superscripts $s$, $v$, and $t$ denote separate parameter sets for the scalar, vector, and tensor channels respectively; $\Psi$ and $\psi$ are MLPs acting on per-node feature norms; and $W^v$ and $W^t$ are learnable linear maps between node and edge channel dimensions (taken as the identity when these dimensions match)

After these, all layers follow a residual update structure, $f'_r =  f_r +  g\left(\|f_r\|\right)\odot M_r(\tilde{f_r};\{\tilde{f_s},\tilde{f_t}\})$, where $g$ is a gating MLP with a sigmoid output that depends on the unnormed features, and $M_r(\tilde{f_r};\{\tilde{f_s},\tilde{f_t}\})$ is a message aggregation with $\tilde{f_r},\tilde{f_s},\tilde{f_t}$
being $\mathrm{E3Norm}$ed receiver, sender, and intermediary features, respectively. In general, 
\begin{equation}
M_r(\tilde{f_r};\{\tilde{f_s},\tilde{f_t}\})=\sum_{s,t\in\mathcal{N}_t(r)}m_{}\left(\tilde{f}_r,\tilde{f}_s,\tilde{f}_{t}\right)\odot\Gamma(\mathbf{R}_r,\mathbf{R}_s)\odot W\tilde{f}_s,
\end{equation}
where $W$ is a linear map to match dimensions (if necessary), and the message generating function is an MLP
\begin{equation}
    m\left(
    \tilde{f}_r,\tilde{f}_s,\tilde{f}_{t}\right)=
    \mathrm{MLP}\left(\tilde{h}_r,\tilde{h}_s,
    \|U\tilde{\mathbf{f}}_r\|,
    \|V\tilde{\mathbf{f}}_s\|,
    \cos(U\tilde{\mathbf{f}}_r,V\tilde{\mathbf{f}}_s),
    \|\tilde{f}_t\|
    \right),
\end{equation}
with $U$ and $V$ being linear maps applied to the $L>0$ channels, and $\cos(A,B)$ being the cosine similarity of the $L>0$ channels. This way, the messages are generated based on both magnitude and directional information. 

The geometric filter $\Gamma$ takes as input an invariant geometric relation between the receiver and sender real-space geometric information, $\mathbf{R}_r,\mathbf{R}_s$, and is absent in layers where there is no obvious geometric relation or to avoid redundant information from previous transformations (e.g., edge-to-node messages in late layers). In node-to-node and node-to-edge messages, we use radial basis functions evaluated at the length of the edge (i.e. distance between node and receiver nodes). In edge-to-edge messages, we project $\cos\left(\mathbf{G}_r,\mathbf{G}_s\right)=\mathrm{Tr}\left(\hat{\mathbf{G}}_r^T\hat{\mathbf{G}}_s\right)$ onto a basis of Legendre polynomials, $\{P_n(x)\}$, since the gyration tensors contain angular information.

At certain steps, geometry is re-injected to avoid degradation of geometric signal. In this case, the residual connection gets two simultaneous updates, one for features and one for geometry,
\begin{equation}
    f'_r = f_r + g_f(\|f_r\|)\odot M_r+g_\mathbf{R}(\|f_r\|)\odot\Delta\mathbf{f}_r,
\end{equation}
where 
\begin{equation}
    \Delta\mathbf{f}_r= \sum_{s,t\in\mathcal{N}_t(r)} m\left(\tilde{f}_r,\tilde{f}_s,\tilde{f}_{t}\right) \odot \Gamma(\mathbf{R}_r,\mathbf{R}_s) \otimes \Delta\hat{\mathbf{R}}_{rs}.
\end{equation}
Here, $\Delta\hat{\mathbf{R}}_{rs}=(\hat{\mathbf{r}}_{rs},\hat{\mathbf{G}}_{rs})$ is a unit-norm tensor. 

At the end of a messaging cycle, we apply a feed-forward residual update following the expression
\begin{equation}
    f_r' = f_r + g(\|f_r\|)\odot 
    \mathrm{MLP}(\|\tilde{f}_r\|, W\cdot\mathrm{EMBED}(Z_r))\odot(\mathbb{I}_0|\tilde{\mathbf{f}}_r)
\end{equation}
where $\mathbb{I}_0$ is the identity over the scalar channels (i.e. scalar features are not gated). We include a linear transformation of the initial embedding of the atomic species whenever this layer is applied to nodes, but that term is omitted when applied to edges.

Finally, the prediction of the model for the properties of a given atom, $\hat{y}_i=(\hat{N}_i,\hat{\lambda}_i,\hat{\boldsymbol{\mu}}_i,\hat{\mathbf{Q}}_i)$ is a readout that depends on an atomic head at the end of each layer:
\begin{equation*}
    (\hat{N}_i^{(l)},\hat{\lambda}_i^{(l)}) = \mathrm{MLP}(h_i^{(l)}), \quad
    \hat{\boldsymbol{\mu}}_i^{(l)} = \sum_cw^{v,(l)}_c\mathbf{x}_{i,c}^{(l)}, \quad
    \hat{\mathbf{Q}}_i^{(l)} = \sum_cw^{t,(l)}_c\mathbf{T}_{i,c}^{(l)},
\end{equation*}
\begin{equation*}
    \text{and} \
    \hat{y}_i = \sum_l \alpha^{(l)} \hat{y}_i^{(l)},
\end{equation*}
where the layer weights $\alpha$ come from a softmax, $\alpha^{(l)} = \exp(\omega^{(l)}) / \sum_{l'}\exp(\omega^{(l')})$.

\subsection{Parity plots for the OOD predictions of QT-Net on AIMEl}

\begin{figure}[h]
    \centering
    \includegraphics[width=0.8\linewidth]{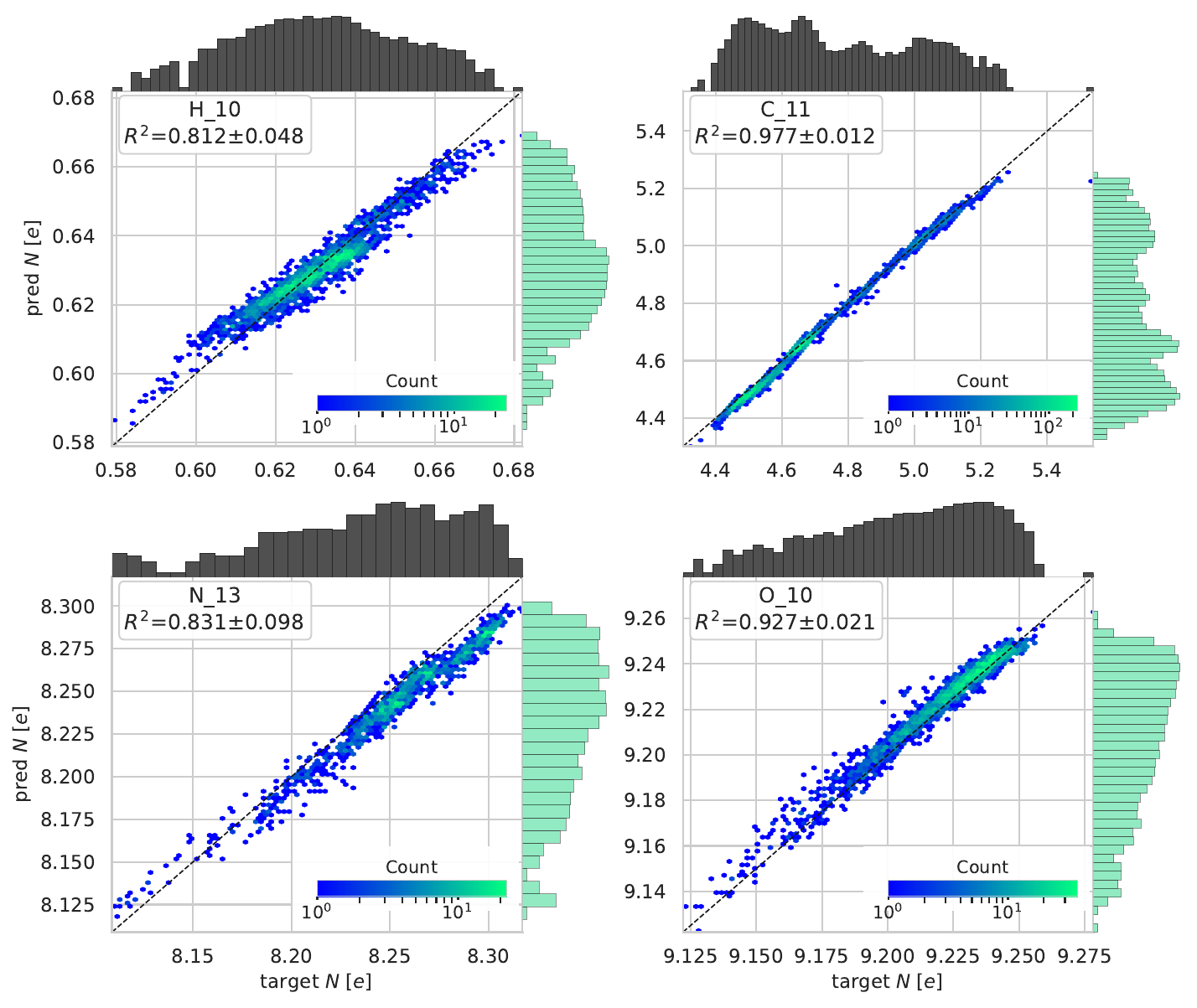}
    \caption{Parity plot for the OOD predictions of QT-Net for electron populations.}
    \label{fig:qtnet_parity_N}
\end{figure}

\begin{figure}[h]
    \centering
    \includegraphics[width=0.8\linewidth]{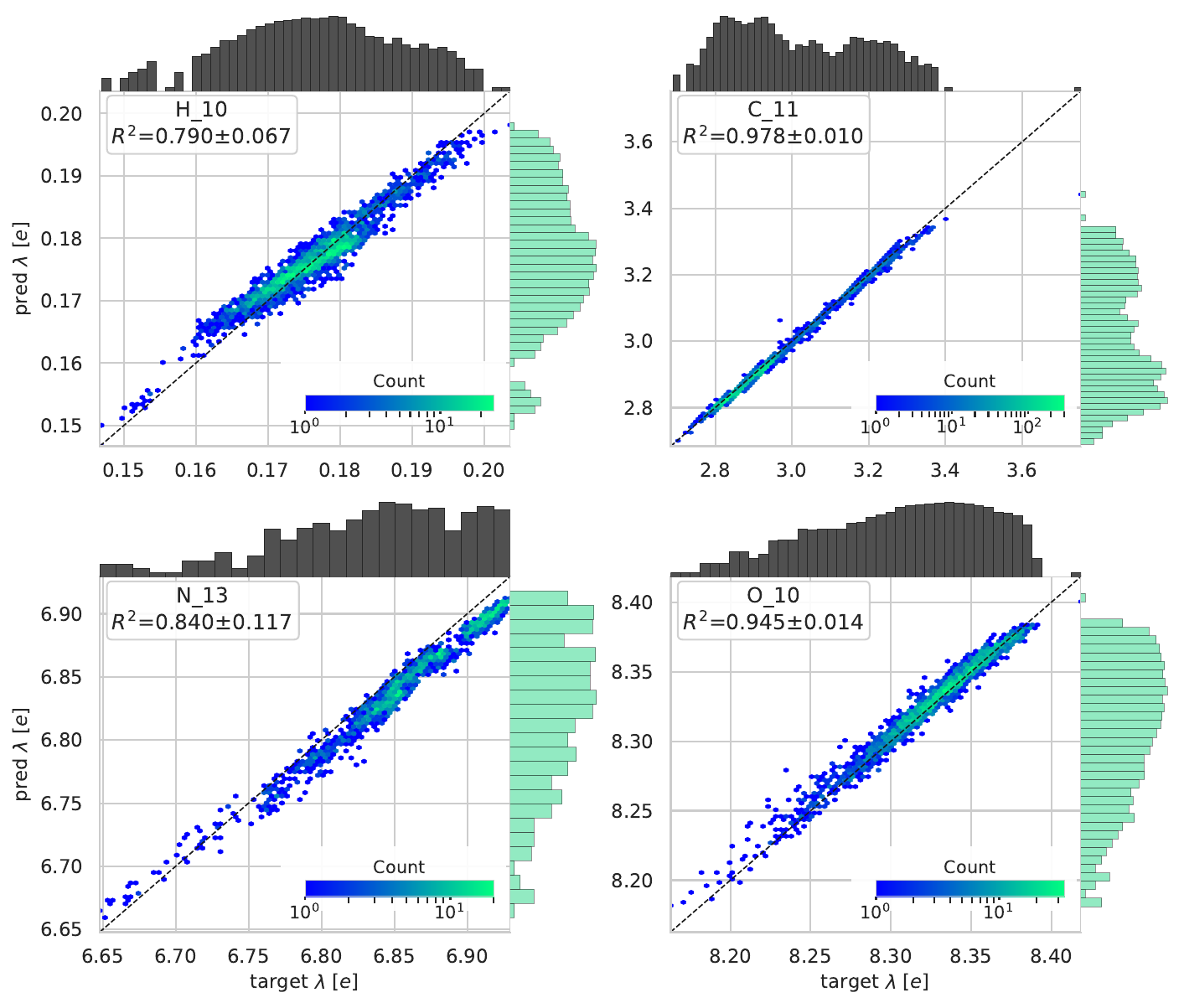}
    \caption{Parity plot for the OOD predictions of QT-Net for localization indices.}
    \label{fig:qtnet_parity_LI}
\end{figure}

\begin{figure}[h]
    \centering
    \includegraphics[width=0.8\linewidth]{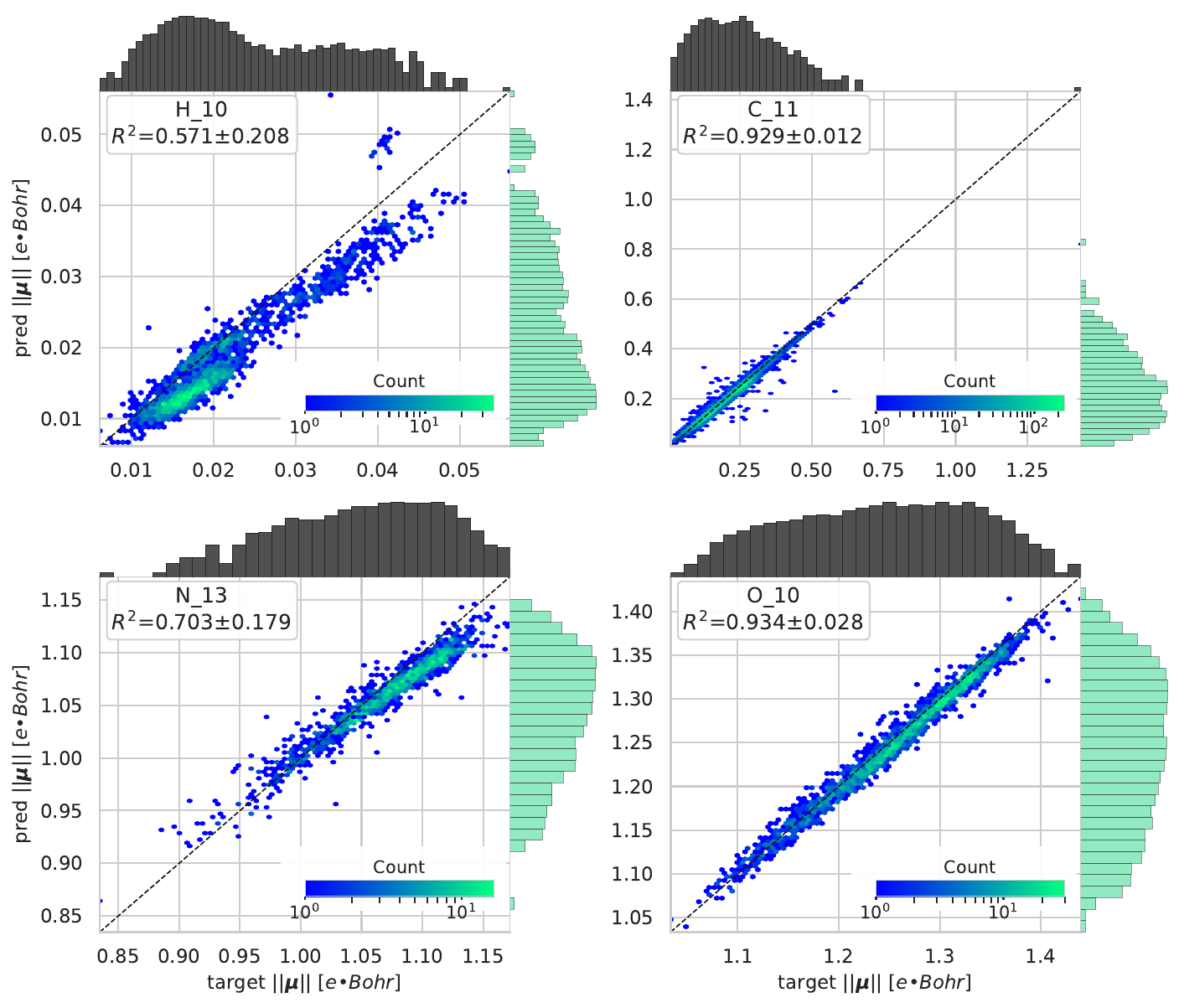}
    \caption{Parity plot for the OOD predictions of QT-Net for atomic contributions to the molecular dipole moment.}
    \label{fig:qtnet_parity_Mu}
\end{figure}

\begin{figure}[h]
    \centering
    \includegraphics[width=0.8\linewidth]{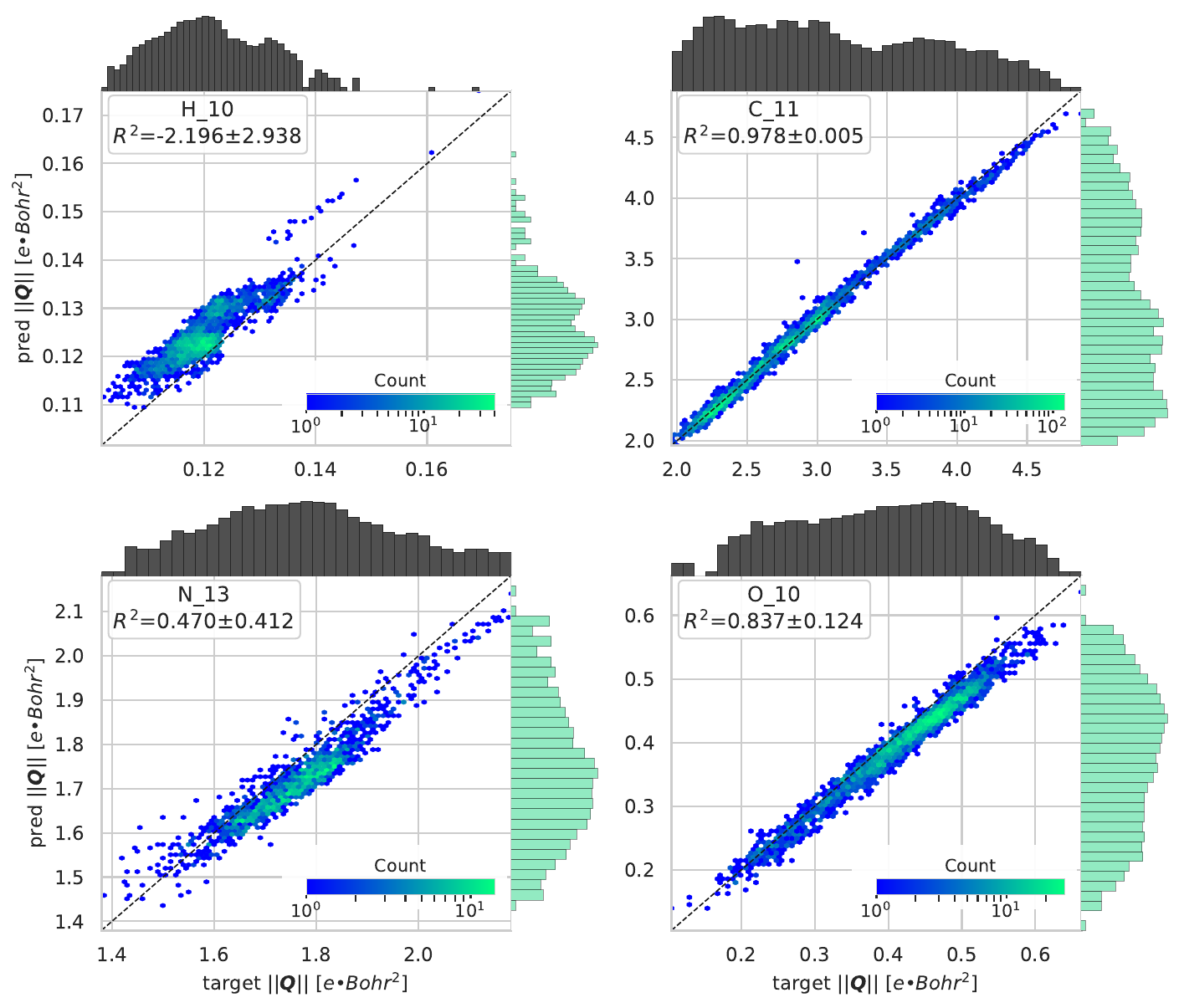}
    \caption{Parity plot for the OOD predictions of QT-Net for atomic quadrupole moments.}
    \label{fig:qtnet_parity_Q}
\end{figure}

We show parity plots for QT-Net's predictions on the held-out atomic environments \{H\_10, C\_11, N\_13, O\_10\} for each of the four QTA targets: electron populations $N$ (Fig.~\ref{fig:qtnet_parity_N}), localization indices $\lambda$ (Fig.~\ref{fig:qtnet_parity_LI}), atomic contributions to the molecular dipole moment $\boldsymbol{\mu}$ (Fig.~\ref{fig:qtnet_parity_Mu}), and quadrupole moments $\mathbf{Q}$ (Fig.~\ref{fig:qtnet_parity_Q}).

\clearpage
\newpage

\end{document}